\RequirePackage{fix-cm}
\documentclass[twocolumn]{svjour3} 

\smartqed
\journalname{International Journal of Computer Vision}

\usepackage{bm}
\usepackage[table]{xcolor}
\usepackage{graphicx}
\usepackage{newtxtext, newtxmath}
\usepackage{balance}
\usepackage{amsmath}
\usepackage[colorlinks=true, linkcolor=blue, citecolor=blue, urlcolor=blue]{hyperref}
\usepackage{xcolor}
\usepackage{latexsym}
\usepackage{makecell}
\usepackage{diagbox}
\usepackage{adjustbox}
\usepackage{caption}
\usepackage{subcaption}
\usepackage{soul}

\usepackage{booktabs}
\usepackage{multirow}
\usepackage{stfloats}
\usepackage{threeparttable}
\usepackage{url}
\usepackage{array}
\usepackage{ragged2e}
\usepackage{enumitem}
\usepackage{microtype}
\usepackage{pifont}

\usepackage{enumerate}

\usepackage[round, sort]{natbib}

\usepackage{booktabs}
\usepackage{multirow}
\usepackage[table,xcdraw]{xcolor}

\usepackage{soul} 
\usepackage[table,xcdraw]{xcolor}
\usepackage{xcolor}

\usepackage{algorithm}
\usepackage{algorithmicx}
\usepackage[noend]{algpseudocode}
\renewcommand{\algorithmicrequire}{\textbf{Input:}}
\renewcommand{\algorithmicensure}{\textbf{Output:}}
\newcommand{\IJCVRevised}[1]{{\color{black}#1}}

\begin{document}
\sloppy

\title{Object-Scene-Camera Decomposition and Recomposition for Data Efficient Monocular 3D Object Detection}

\author{Zhaonian Kuang \and
        Rui Ding \and
        Meng Yang \and 
        Xinhu Zheng \and
        Gang Hua \and
}

\institute{ Zhaonian Kuang \and Rui Ding \and Meng Yang \at
  Institute of Artificial Intelligence and Robotics, Xi'an Jiaotong University, Xi'an, P.R.China \\ 
  \email{znkwong@stu.xjtu.edu.cn,\\ dingrui@stu.xjtu.edu.cn,\\ mengyang@mail.xjtu.edu.cn}
           \and
  Xinhu Zheng \at
  Intelligent Transportation Thrust of the Systems Hub, Hong Kong University of Science and Technology (GZ), Guangzhou, P.R.China \\
  \email{xinhuzheng@hkust-gz.edu.cn}
           \and           
           Gang Hua \at
  Amazon Alexa AI, Bellevue, Washington, United States \\
  \email{ganghua@gmail.com}
            \and
    Codes and models are available at \href{https://github.com/kwong292521/DR-Traversal-M3D}{\textcolor{blue}{this website}}.
}

\date{Received: 22 July 2025}

\maketitle
\begin{abstract}

Monocular 3D object detection (M3OD) is intrinsically ill-posed, hence training a high-performance deep learning based M3OD model requires a humongous amount of labeled data with complicated visual variation from diverse scenes, variety of objects and camera poses.
However, we observe that, due to strong human bias, the three independent entities, i.e., object, scene, and camera pose, are always tightly entangled when an image is captured to construct training data. 
More specifically, specific 3D objects are always captured in particular scenes with fixed camera poses, and hence lacks necessary diversity. 
Such tight entanglement induces the challenging issues of insufficient utilization and overfitting to uniform training data. 
To mitigate this, we propose an online object-scene-camera decomposition and recomposition data manipulation scheme to more efficiently exploit the training data. 
We first fully decompose training images into textured 3D object point models and background scenes in an efficient computation and storage manner. 
We then continuously recompose new training images in each epoch by inserting the 3D objects into the freespace of the background scenes, and rendering them with perturbed camera poses from textured 3D point representation.
In this way, the refreshed training data in all epochs can cover the full spectrum of independent object, scene, and camera pose combinations.
This scheme can serve as a plug-and-play component to boost M3OD models, working flexibly with both fully and sparsely supervised settings. In the fully-supervised setting, all objects are annotated. In the sparsely-supervised setting, objects closest to the ego-camera for all instances are sparsely annotated. We then can flexibly increase the annotated objects to control annotation cost. 
For validation, our method is widely applied to \IJCVRevised{five} representative M3OD models and evaluated on both the KITTI and the more complicated Waymo datasets. In the fully-supervised setting, our method significantly improves the performance of the base models by 26\%$\sim$48\% relatively, 
and achieves new state-of-the-art on KITTI as of submission. In the sparsely-supervised setting, our method with only 10\% annotations achieves on-par performance with fully-supervised paradigm in most scenarios.

\end{abstract}


\begin{figure}[t]
\centering
\includegraphics[width=1.0\columnwidth]{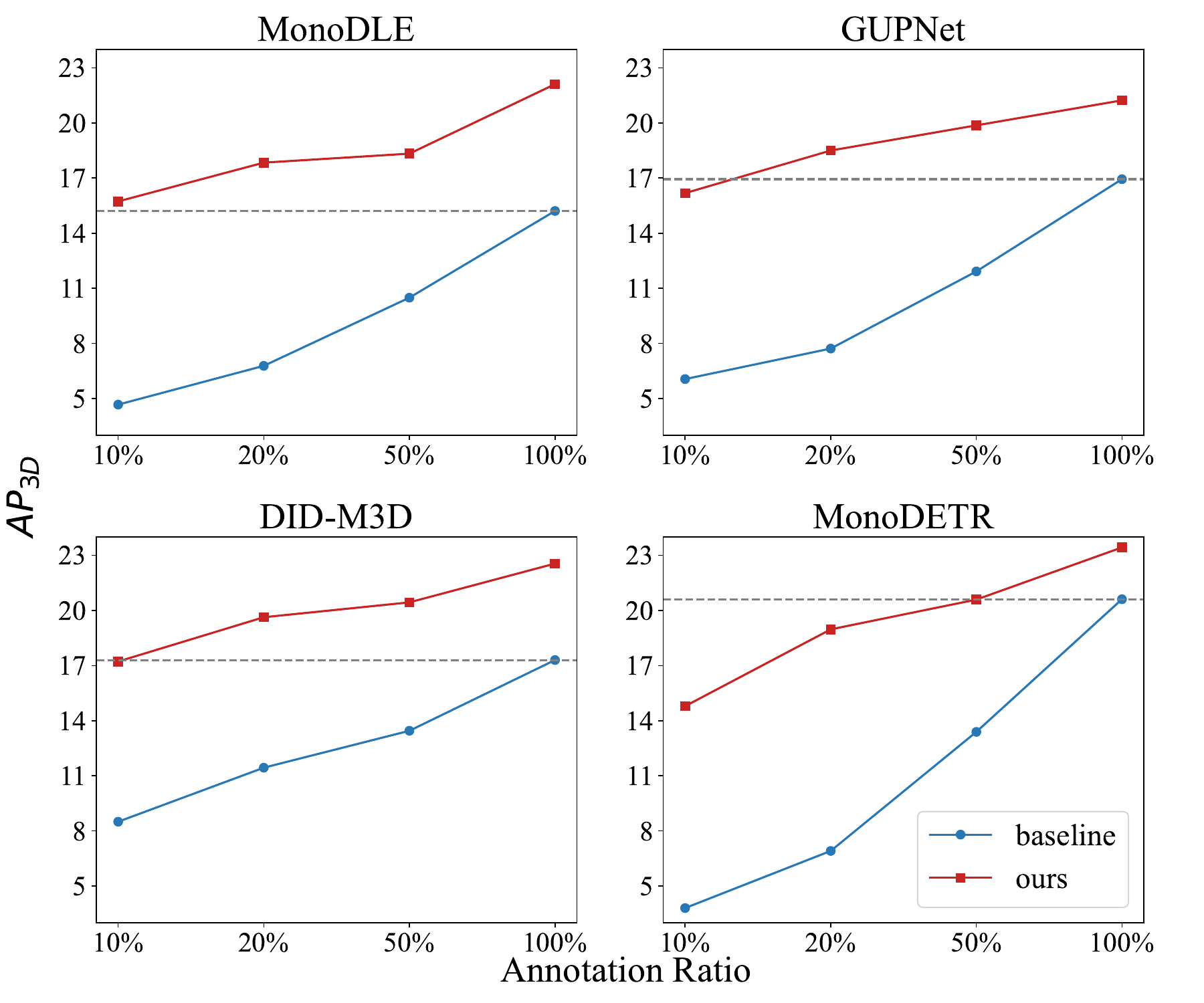}
\caption{Implementation of our data manipulation scheme on four released models under both fully and sparsely supervised settings.
The $x$-axis represents the percentage of annotated objects used in our scheme out of all objects in the KITTI train split. The $y$-axis represents the moderate level of $AP_{3D}$ in the KITTI validation split. Our method significantly boosts the performance of the base models in both the fully (i.e., 100\% annotation ratio) and the sparsely (i.e., 10\%, 20\%, and 50\%) supervised settings. Our method with only 10\% annotations even achieves on-par performance with the fully-supervised setting of the first three base models. 
}
\label{fig_label_ratio}
\end{figure}

\section{Introduction}
Monocular 3D object detection (M3OD)  
~\citep{ma2021delving,reading2021categorical,lu2021geometry,peng2022did,zhang2023monodetr,huang2023obmo} 
aims to restore the 3D attributes of objects in a scene, such as 3D position, orientation, and size, using a single RGB image. Compared with high-cost LiDAR-based schemes ~\citep{lang2019pointpillars,shi2023pv,zhang2024comprehensive,sheng2025ct3d++,zhang2025cmae,yang2025fusion4dal} and multi-view schemes ~\citep{li2023bevdepth,li2022bevformer,wang2025physically,wu2024heightformer,huang2021bevdet,zhang2022beverse,huang2025lidar}, M3OD provides a low-cost alternative for autonomous driving and robot navigation. However, M3OD is more challenging due to the intrinsic ill-posedness of monocular 3D vision.

Training a high-performance deep learning based M3OD model requires a humongous amount of labeled data with complicated visual variation from diverse scenes, a variety of objects, and camera poses in real world. This could help the network learn a robust feature representation to alleviate the ill-posed problem.
However, the annotations of 3D object detection are more challenging and costly compared with conventional 2D object detection. 
Limited training data may result in low accuracy of M3OD models and difficult application in real world.
Therefore, developing a data efficient scheme is a meaningful yet challenging task for M3OD. 

Existing work has made several efforts to alleviate the training data problem.  
Conventional image-level augmentations ~\citep{liu2022learning} are basic schemes. Semi-supervised schemes ~\citep{yang2023mix, zhang2024decoupled} directly alleviate the problem by scaling up training data but is limited by the quality of pseudo labels.
Copy-Paste ~\citep{lian2022exploring, qiao2024monosample} can generate more objects with labels for M3OD. It achieves this by cutting and pasting the image patches of objects. However, these schemes are limited by inserting positions in the scene due to the fixed occupation of existing objects. Moreover, the 2D-3D geometric consistency after paste cannot be maintained.
Recent Nerf-based method ~\citep{tong20233d}, GAN-based ~\citep{li2023lift3d}, graphical-engine-based method ~\citep{ge20233d} and diffusion-based method ~\citep{wen2024panacea, parihar2025monoplace3d} can generate precise 3D data but suffer from high reconstruction or rendering cost, making them difficult to embed into the training pipeline of M3OD models. 

In addition, camera pose is an important factor of training data for M3OD, but is often overlooked. 
In traffic scenes, camera pose is entangled with the ground plane prior ~\citep{dijk2019neural, zhou2021monoef}, the latter of which is often utilized by M3OD models to estimate the object's depth. Such tight entanglement makes the model sensitive to camera pose perturbation.
MonoEF ~\citep{zhou2021monoef} improves the robustness to pose changes by learning and rectifying camera pose online, however, it does not directly reduce pose sensitivity. Moreover, training data collected with a fixed camera pose are typically annotated at a low temporal resolution. This setup fails to capture the full data stream, limiting the model capacity to learn from the training data.

In this paper, we observe that one crucial problem of training data for M3OD lies in the tight entanglement of the three entities, i.e. object, scene, and camera pose.
That is, to construct training samples, specific 3D objects are always captured in particular scenes with fixed camera poses. Due to this strong human bias, the training data lacks necessary diversity.
However, these three entities are essentially independent in the real world.
For example, the traffic scene and the camera pose are diverse in real scenarios, and different objects can appear anywhere in the freespace of a scene. 
This tight entanglement problem appears everywhere in real-world datasets for M3OD such as the KITTI and the more complicated Waymo. 
It results in the network being trained for multiple epochs on uniform training data, where objects, scenes, and camera poses are tightly entangled. 

This will results in three challenging issues: 

\textbf{(i) Overfitting to uniform training data.} 
To estimate 3D properties, especially depth, the detector relies on: (1) the appearance and 2D size of the object; (2) the relationship between the object and the scene; (3) the relationship between adjacent objects. Due to the tight entanglement of objects and scenes, the appearance, 2D size of the object, object-scene relationship and object-object relationship are remain the same in each training epoch. Therefore, the network is prone to overfitting to these uniform training data. 

\textbf{(ii) Underutilization of object-scene and object-object relationship.} 
Objects and scenes are supposed to be two independent entities in real world. Due to their tight entanglement, objects can only appear on the specific locations of specific scenes. 
Firstly, the fixed object appearance and 2D size hinder the network from learning robust semantic-to-3D properties mapping. 
Secondly, the fixed location of the object hinders the network to exploit more structural information from the scene, which helps estimate the object's depth.  
Thirdly, the spatial relationships between adjacent objects can also assist in depth estimation, but a fixed object-object relationship hinders the model from learning a robust depth reasoning pattern based on this relationship.

\textbf{(iii) Limited camera pose variation}.
The ground plane prior in the traffic scene ~\citep{dijk2019neural, zhou2021monoef} is tightly entangled to camera pose. As a result, when the network learns these priors, it becomes sensitive to specific pose perturbations due to the tight entanglement between 3D scenes and camera poses.
In addition, collected training data with a fixed camera pose is typically annotated at a low temporal resolution, which limits the model's ability to learn from the training data effectively.

Existing methods address these issues to some extent. For example, Copy-Paste ~\citep{lian2022exploring, qiao2024monosample} can extend the object-scene and object-object relationship, but cannot alleviate the two issues of overfitting to uniform training data and limited camera pose variation. Recent reconstruction and generation methods ~\citep{tong20233d, li2023lift3d, ge20233d, wen2024panacea} have the potential to solve these three issues, but face prohibitive computational costs for M3OD tasks. Specifically, the augmented data construction process of these methods have to be deployed offline. It requires substantial GPU hours for computation and a large amount of disk space for data storage. 

To mitigate the three issues, we propose an online object-scene-camera decomposition and recomposition data manipulation scheme 
for data efficient M3OD.
Our scheme fully decomposes and randomly recomposes the three independent entities in 3D space, i.e., objects, scenes, and camera poses, to exploit training data more efficiently.
In our scheme, object-scene decomposition helps the network alleviate the overfitting to uniform training data, object-scene recomposition helps explore more object-scene and object-object relationships, and scene-camera decomposition and recomposition can extend the camera pose of a frame from fixed value to diverse variations.
 
More specifically, in the decomposition process, we construct objects as textured 3D point models, and remove all objects from the scene to construct empty scenes. In this process, we utilize texture point cloud representation for efficient computation and storage.
In the recomposition process, we continuously recompose new training images online in each epoch by traversing the combination of objects, scenes, and camera poses. 3D objects are inserted into the freespace of scenes to explore variation of object appearance and 2D size, and the relationships of objects and scenes. Moreover, the 3D texture scene point cloud is rendered into 2D images with perturbed camera poses. 
This way, the refreshed training data in all epochs can cover the full spectrum of independent object, scene, and camera combinations.
Thus, the network can learn robust representations related to all objects, scenes, and camera poses in 3D scenes.

Our scheme can serve as a plug-and-play component to boost deep learning based M3OD models, and it works flexibly with both fully and sparsely supervised settings. In the fully-supervised setting, all objects are annotated to apply our scheme. In the sparsely-supervised setting, only the objects closest to the ego-camera for all instances are sparsely annotated to avoid redundant semantics across successive frames. Then we can flexibly increase the number of annotated objects to control annotation costs for easier applications in real world. 

To validate our scheme, we widely apply it to \IJCVRevised{five} representative M3OD models with publicly released codes, including MonoDLE ~\citep{ma2021delving}, GUPNet ~\citep{lu2021geometry}, DID-M3D ~\citep{peng2022did}, MonoDETR ~\citep{zhang2023monodetr}, \IJCVRevised{MonoDGP ~\citep{pu2025monodgp}}, and evaluate them on both the widely used KITTI and the more complicated Waymo benchmarks. \IJCVRevised{We also apply our scheme to PETR ~\citep{liu2022petr} under multi-camera setting on Waymo.}

In the fully-supervised setting, our scheme significantly improves $AP_{3D}$ of the four base models by 26\%$\sim$48\% relative to their original performance, and achieves new state-of-the-art (SOTA) on KITTI.
In the sparsely-supervised setting, our method with fewer annotations (especially only 10\% for MonoDLE, GUPNet and DID-M3D) can achieve on-par performance, when compared to fully-supervised paradigm of these base models. The results in these two settings are shown in Figure \ref{fig_label_ratio}.
Notably, our method holds great potential to achieve higher accuracy with stronger base models upon their codes released.

We highlight the main contributions as follows:

\begin{itemize}
    \item We observe the crucial tight entanglement problem of the three independent entities,i.e., scene, object, and camera pose, seriously hinders the efficient utilization of training data in M3OD.
    \item We propose an online data manipulation scheme to mitigate the issues of insufficient utilization and overfitting on uniform training data. It serves as a plug-and-play component to boost M3OD models, working flexibly with both fully and sparsely supervised settings. 
    \item Our scheme significantly improves the performance of recent M3OD models and achieves SOTA on KITTI. In addition, our scheme with fewer annotations can achieve on-par performance with fully-supervised paradigm.
\end{itemize}
\section{Related Work}

\subsection{Monocular 3D Object Detection}
Monocular 3D object detection (M3OD) focuses on detecting and localizing objects in 3D space based on a single RGB image. Most M3OD methods are developed based on traditional 2D object detection methods ~\citep{ren2015faster, tian2019fcos, zhou2019objects}. However, compared to 2D Detection, ~\citep{ma2021delving} and ~\citep{wang2022probabilistic} reveal that depth estimation is the key challenge in M3OD.

To tackle the challenging problem of accurately estimating depth from ill-posed monocular 3D vision, various methods have been proposed, each employing a distinct approaches. MonoDLE and MonoCon~\citep{ma2021delving, liu2022learning} estimate object depth through direct regression. Additionally, MonoCon ~\citep{liu2022learning} advocates learning auxiliary monocular contexts via keypoint estimation.
MonoPair ~\citep{chen2020monopair} utilizes the spatial relationship of objects to help estimating depth of per objects.
CaDDN ~\citep{reading2021categorical} uses categorical depth distribution estimation to elevate 2D features into 3D features. 
GUPNet ~\citep{lu2021geometry} combines 2D height and 3D height of objects with perspective projection to determine object depth.
DID-M3D ~\citep{peng2022did} decouples object depth into surface depth and attribute depth.
MonoLSS ~\citep{li2024monolss} uses learnable ROI feature selection to help estimate depth.
MonoFlex ~\citep{zhang2021objects}, MonoDDE ~\citep{li2022diversity}, and MonoCD ~\citep{yan2024monocd} use the 2D-3D geometric constraint of object keypoints for depth inference.
MonoDETR ~\citep{zhang2023monodetr} and MonoDTR ~\citep{huang2022monodtr} introduce the transformer architecture for global relationship modeling to strengthen depth prediction.
MonoEF ~\citep{zhou2021monoef} uses online feature rectification to make model robust to camera pose perturbations.

To effectively alleviate the ill-posed depth estimation problem, a huge amount of labeled data is required for training. However, the annotations of M3OD are
much more challenging and costly compared to 2D object detection. Limited training data may lead to low accuracy of M3OD models and hinder their application in real-world scenarios. Our work aims to develop a data manipulation scheme for M3OD to exploit training data more efficiently.

\subsection{Boosting M3OD from a Data Pespective}
To mitigate the limitation of training data, many works seeks to enhance monocular 3D detectors from a data perspective. We categorize them into four categories: image-level augmentation, utilizing additional data, copy-paste strategy, and data reconstruction and generation.

\subsubsection{Image-level Augmentation}
Color augmentation, random flipping, and random cropping ~\citep{liu2022learning} are representative image-level augmentation operations. DID-M3D ~\citep{peng2022did} and EGC ~\citep{lian2022exploring} use scaling factors to adjust the depth labels of objects to alleviate the 2D-3D geometric inconsistency caused by image scaling. However, image-level augmentation still suffers from the tight entanglement issue.

\subsubsection{Utilizing Unlabeled Data}
LPCG ~\citep{peng2022lidar}, Mix-Teaching ~\citep{yang2023mix}, and DPL ~\citep{zhang2024decoupled} utilize large-scale unlabeled data, employing pseudo-label mining strategies to expand training samples. However, these methods still suffer from the tight entanglement problem and low-quality pseudo labels.

\subsubsection{Copy-Paste Strategy}
Copy-paste is an effective technique to increase the number of training samples by extracting image patches of certain objects and pasting them into different areas. This technique was first applied to 2D vision tasks ~\citep{dvornik2018modeling, wang2019data, fang2019instaboost}. EGC ~\citep{lian2022exploring} and MonoSample ~\citep{qiao2024monosample} extend this strategy to M3OD by adjusting the 2D size of objects and enforcing object-ground attachment constraints. However, this approach suffers from the 2D-3D inconsistency problem. Morever, the presence of existing objects in the scene restricts the available insertion positions for new objects.
Therefore, while this strategy expands object-scene relationships, it fails to mitigate the issues of overfitting to invariant training data and limited camera pose variation.

\subsubsection{Data Reconstruction and Generation}

3DA ~\citep{tong20233d} and Lift3D ~\citep{li2023lift3d} employ NeRF and GAN, respectively, to generate precise 3D data. ~\citep{ge20233d} employs a graphical engine, while ~\citep{wen2024panacea,parihar2025monoplace3d} utilize a diffusion model to generate more realistic training samples. However, these approaches incur high computational costs for reconstruction and rendering. Therefore, they have to be deployed offline. Consequently, generating a large-scale dataset with diverse object-scene-camera combinations for training remains computationally expensive and impractical.

In summary, existing methods fail to effectively address the tight entanglement issue of object, scene, and camera poses, or come with prohibitive computational costs. In contrast, we conduct a comprehensive investigation and propose an effective solution to this problem. Compared to previous methods, our approach simultaneously achieves 2D-3D consistent data manipulation, covers the full spectrum of object-scene-camera combinations during training while maintains a low computational cost.
\begin{figure*}[t]
\centering
\includegraphics[width=0.95\textwidth]{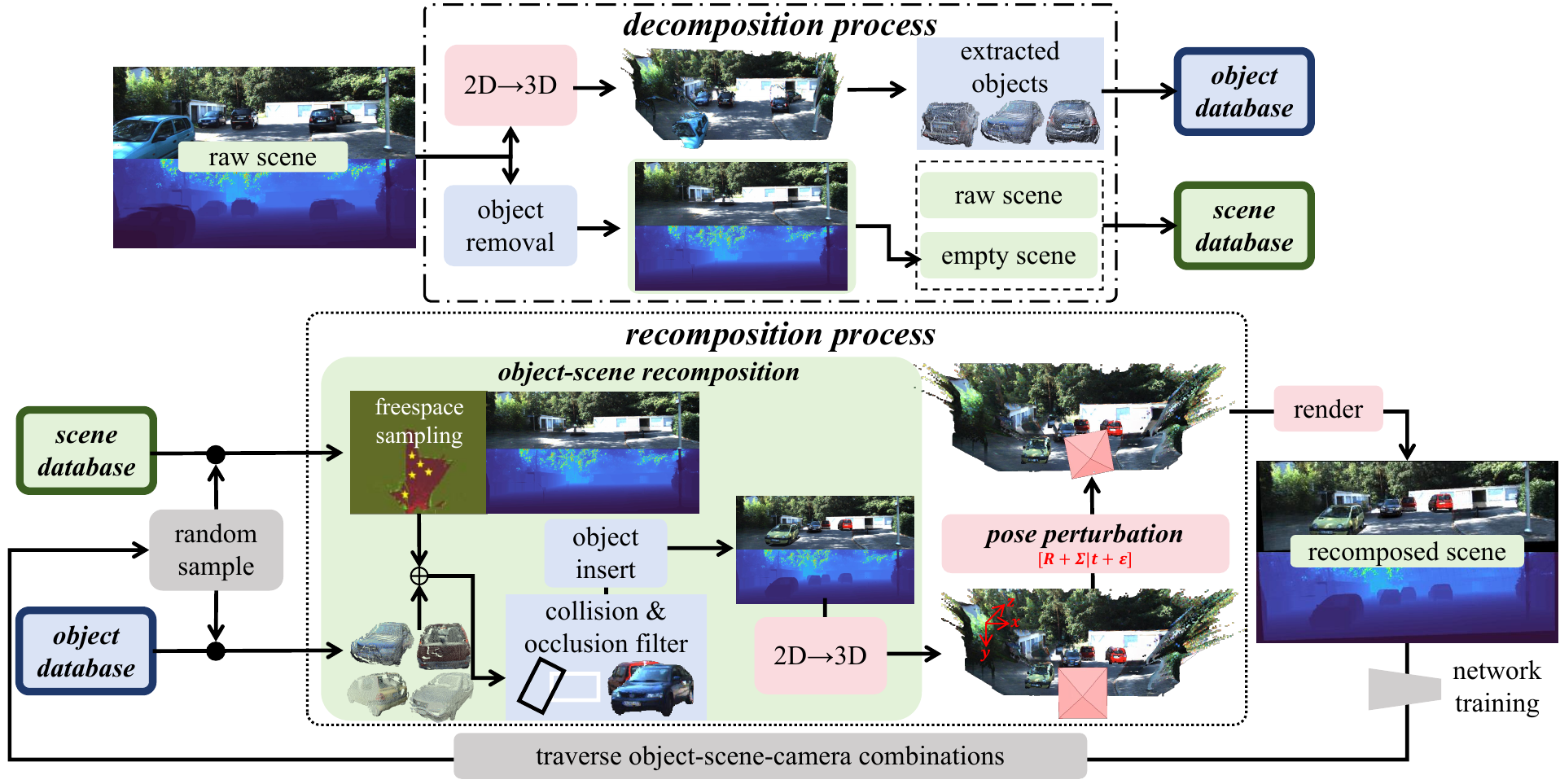} 
\caption{Illustration of our online object-scene-camera decomposition and recomposition data manipulation scheme. 
We first fully decompose training images into textured 3D object point models and background scenes in an efficient computation and storage manner. We then continuously recompose new training images in each epoch by inserting the 3D objects into the freespace of the background scenes, and rendering them with perturbed camera poses from textured 3D point representation. In this way, the refreshed training data in all epochs can cover the full spectrum of independent object, scene, and camera combinations.
}
\label{fig_framework_overview}
\end{figure*}

\section{Methodology}
\subsection{Scheme Overview}
Figure \ref{fig_framework_overview} illustrates our online object-scene-camera decomposition and recomposition data manipulation scheme. It has two main processes, i.e., decomposition and recomposition. 

In the decomposition process, we first reconstruct objects from training images to textured 3D object point models. We then remove all objects from the scenes to construct empty scenes. All object models are collected to construct an object database. All raw and empty scenes are collected to build a scene database. 
Notably, this process is offline deployed and the computation and storage is efficient due to the simple operation.

In the recomposition process, we first randomly sample locations in the scene freespace. We then randomly sample objects from the object database and replace them at the sampled locations after collision and occlusion filters. In this way, we can fully exploit the object information, the object-scene relationship and object-object relationship. We then perturb the camera pose and render recomposed scenes to new training images to alleviate the limited camera pose variation issue.
The recomposition process is computationally efficient and can be embedded into the training pipeline. Therefore, it continuously feeds refreshed data to the model without pre-generation before training.  In this way, the refreshed training data in all epochs can cover the full spectrum of independent object, scene, and camera combinations.

First of all, we give a brief introduction to the data we used. For a raw scene, the data contains an RGB image $I_s$, the synchronous LiDAR point clouds $P_l $ ~\citep{geiger2012we} and the ground plane equation $G : ax+by+cz+d=0$. The ground plane equation can be estimated by RANSAC ~\citep{fang2021LiDAR}, which is applied to the extracted ground LiDAR points. To generate the RGB-Depth scene representation, we use an off-the-shelf depth completion model ~\citep{wang2023g2} to obtain a dense depth map $D_s$ for the scene. Each scene has corresponding camera intrinsic $K$. $T_{2d \xrightarrow{}3d}(D,K,M)$ denotes transforming the 2D depth to 3D points cloud, where $M$ is the foreground mask. We only transform the foreground pixel if $M$ is given. Otherwise, the whole image will be transformed. $T_{3d \xrightarrow{}2d}(P,K)$ denotes transforming the points cloud under camera coordinate system to image space. The 3D coordinate system we use is the camera coordinate system specified by KITTI ~\citep{geiger2012we}. In this system, the $x$-axis points right, the $y$-axis points up and the $z$-axis points forward. We use the u-v coordinate system in the image space.

\begin{figure}[]
\centering
\includegraphics[width=1.0\columnwidth]{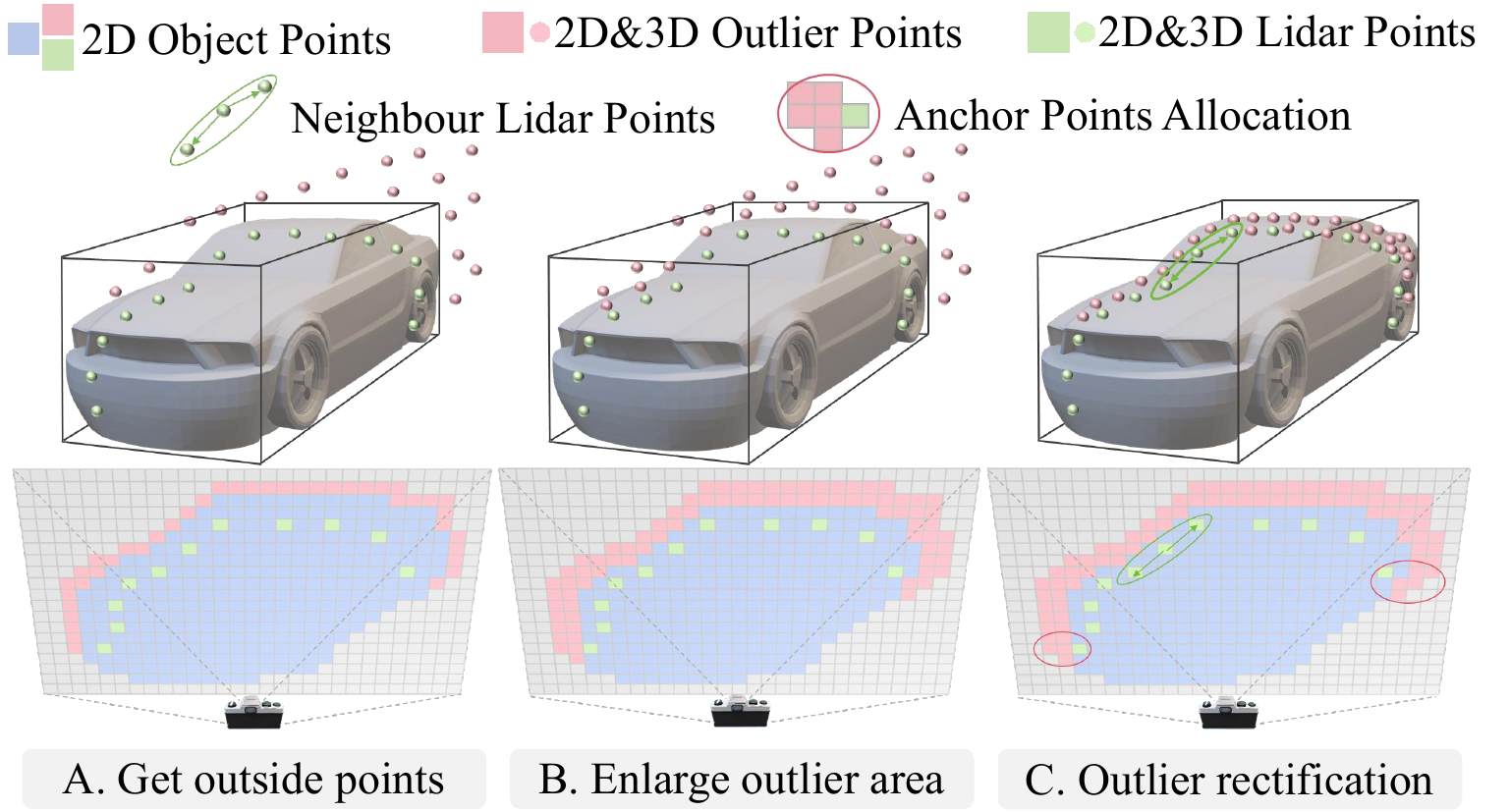}
\caption{Object point model rectification. 
The CAD model is intended to aid visualization.
}
\label{fig_object_rectify}
\end{figure}

\subsection{Decomposition Process}

\subsubsection{Object database construction}  
We aim to construct textured 3D object models using points representation. Compared to copy-paste methods that use 2D image patches ~\citep{lian2022exploring, qiao2024monosample}, this representation can perform more flexible 3D manipulation. It also maintaining 2D-3D geometric consistency. Compared to recent reconstruction and generation-based methods~\citep{tong20233d, li2023lift3d, ge20233d, wen2024panacea}, this representation has a low reconstruction and rendering cost, which is beneficial to our online recomposition process.

To construct the object database, we first use a 2D segmentation method ~\citep{kirillov2023segment} to obtain the foreground masks $M$ of objects. We then reconstruct the 3D points model $\textbf{P}_{og}=T_{2d \xrightarrow{}3d}(D_s, K, M)$, where $\textbf{P}_{og}  \in \mathbb{R}^{k \times 3}$ and $k$ denotes the number of foreground points, together with the texture $\textbf{P}_{oc} \in \mathbb{R}^{k \times 3}$ extracted from $I_s$ using $M$. Finally, we construct the textured 3D points model $\{\textbf{P}_{og}, \textbf{P}_{oc}\}$ and add it to the object database.

Due to the common edge distortion issue caused by the depth completion model, if we directly use the raw completed depth map $D_s$ to construct the objects model, the object model will be harmful to the recomposition process. Hence, we have to rectify the edge depth around the foreground pixels to obtain rectified dense depth $D_{rect}$ and use $D_{rect}$ for object model construction, with $\textbf{P}_{og}^{rect}=T_{2d \xrightarrow{}3d}(D_{rect}, K, M)$.

To solve this issue, we first use an annotated 3D box to identify which points of $\textbf{P}_{og}$ are outside the box. Then these points will be identified as outlier points. We can then obtain the outlier mask $M_{outlier}$ in image space according to the 2D-3D correspondence. Because the surface of the object is continuous, the distorted area is supposed to be larger than the area determined by the outside points. Therefore, we need to enlarge the outlier area by applying a morphological dilation operation to obtain new $M_{outlier}$. Once the outlier area is refined, we conduct rectification on each outlier point.

Specifically, for each outlier point, we find the nearest LiDAR points in image space as anchor points. Then, we use the rectified scale of the anchor points to rectify the depth of the outlier point. To determine the rectified scale, for each LiDAR point, we find a few neighbor LiDAR points nearest to it in 3D space and compute the rectified scale $s_{anchor}$:

\begin{equation}
    s_{anchor}=\frac{1}{n_{nb}}\sum_{i=1}^{n_{nb}}|z_{anchor}-z_{neighbour}^{i}|,
    \label{eq:s_anchor}
\end{equation}
where $z_{anchor}$ and $z_{neighbour}$ are the depth of the anchor LiDAR point and the neighbor LiDAR points, and $n_{nb}$ is the number of neighbor points to be selected. We then use Equation (\ref{eq:z_rect}) to rectify the depth of each outlier point:

\begin{equation}
    z_{rect} = (\frac{2}{1+e^{-z_{outlier}}}-1) * s_{anchor} + z_{anchor}.
    \label{eq:z_rect}
\end{equation}

\begin{figure}[]
\centering
\includegraphics[width=1.0\columnwidth]{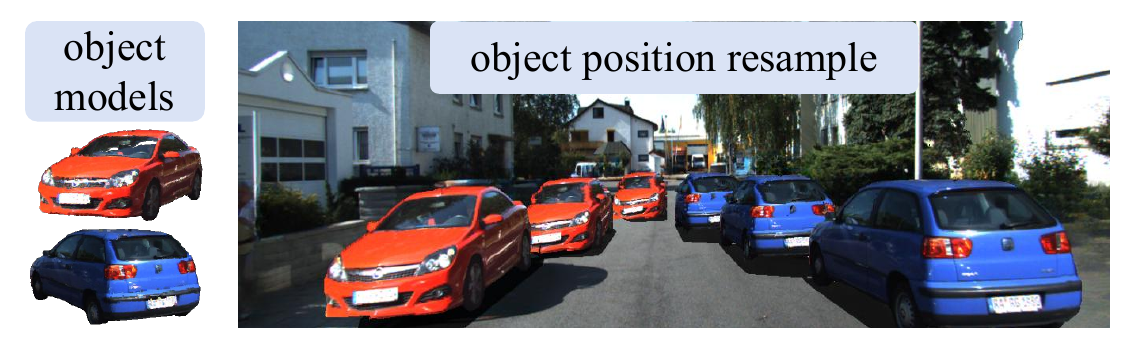}
\caption{Object position resampling. Using a 3D textured point representation, objects can be flexibly resampled to new positions.
}
\label{fig_object_resample}
\end{figure}

This process is shown in Figure \ref{fig_object_rectify}. Based on the textured 3D object point model, we can flexibly resample objects to new positions, as shown in Figure \ref{fig_object_resample}.

\subsubsection{Scene database construction} 
A raw scene is often occupied by fixed objects, which can cause the network to overfit on duplicated data and hinder the exploitation of the object-scene relationship. To solve this, we remove all objects from the raw scene and create an empty scene in RGB-Depth representation. The images and depth maps of the scene will be processed separately. For the image, we use an object removal method, LaMa ~\citep{suvorov2022resolution}, to eliminate the semantics of objects, obtaining an empty scene image $I_{es}$. For the depth map, we use ground depth and background depth to replace the foreground depth. 

\begin{figure}[]
\centering
\includegraphics[width=1.0\columnwidth]{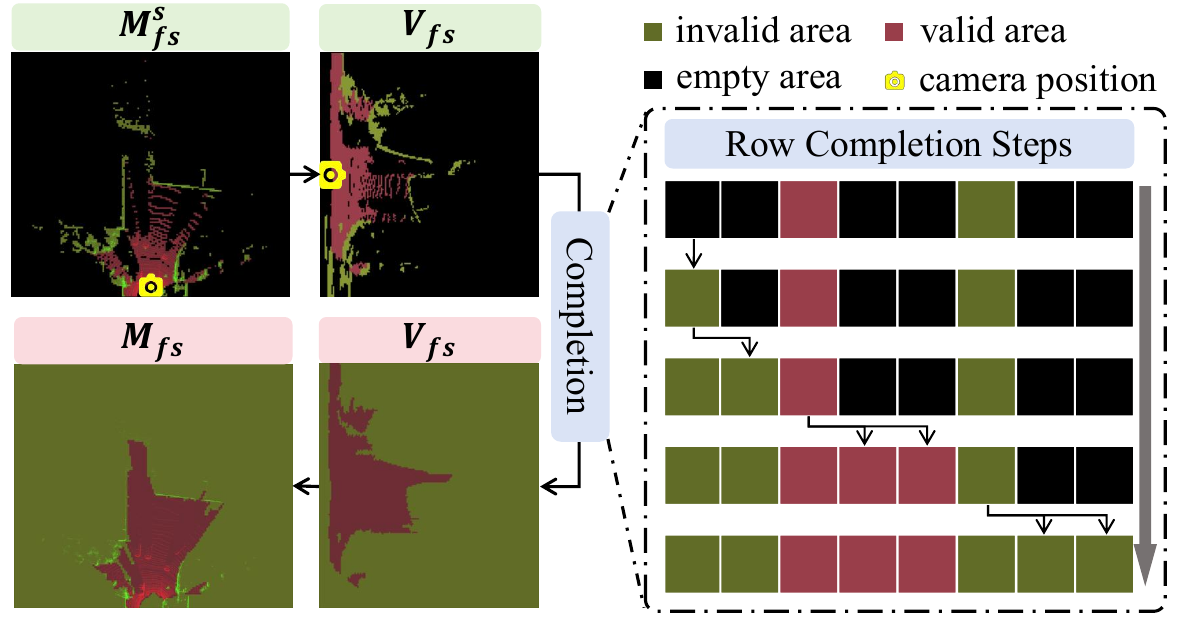} 
\caption{Freespace generation process. 
For each row of the freespace map in the polar coordinates system, we scan each column to complete the empty area.
}
\label{fig_freespace_gen}
\end{figure}

Specifically, to generate the background depth map $D_{bg}$, we first copy it as $D_{bg}=D_{s}$. Then, for each column of pixels in the foreground area, we allocate their depth value as the depth value of the upper neighbor of the top pixel among these column pixels. This can be formulated as:

\begin{equation}
    D_{bg}^{(i,j)} = D_{bg}^{(i, k)}, k=min(\{v | (i,v) \in M_{fg}\})-1
    \label{eq:D_bg},
\end{equation}
where $M_{fg}$ is the foreground area. To generate the ground depth map $D_{ground}$, we use the ground plane equation $G$ of the scene to render the depth, setting non-ground areas to infinite depth value. Finally, we copy the empty scene depth as $D_{es}=D_{s}$, and apply pixel-wise minimum in the foreground area between $D_{bg}$ and $D_{ground}$ to replace the corresponding pixels of $D_{es}$:

\begin{equation}
    D_{es}[M_{fg}] = min(D_{bg}[M_{fg}], D_{ground}[M_{fg}]).
    \label{D_es}
\end{equation}

After that, we save the empty scene $(I_{es}, D_{es})$ and the raw scene $(I_{s}, D_{s})$ to the scene database. 

\subsubsection{Scene freespace generation}\label{sec:Scene freespace generation}
The scene space is vast. Apart from areas occupied by static obstacles such as houses and trees, objects can appear in any other areas of the scene, which belong to the freespace of the scene. We identify freespace using LiDAR data $P_{l}$. 
Initially, LiDAR points of existing objects are projected onto the ground using the ground-plane equation $G$. We then use the method in LiDAR-aug ~\citep{fang2021LiDAR} to generate sparse freespace map $M_{fs}^s \in \mathbb{R}^{h \times w}$ in BEV space, where each grid represents a $r_{fs} \times r_{fs}$ area, as shown in Figure \ref{fig_freespace_gen}.

In the sparse freespace map $M_{fs}^s$ , ground plane belongs to valid areas, and invalid areas are static obstacles. However, due to the sparsity of LiDAR data, there are many empty areas where there are no LiDAR points. These empty areas remain unclassified, which limits object-scene recomposition. To address this, we complete the empty areas with assistance from adjacent valid or invalid regions.

Considering the characteristics of LiDAR scans, completing the freespace map is more effective in the Polar coordinates system. Hence, we transform the freespace map $M_{fs}^s$ from the Cartesian to the Polar coordinates system centered at the camera position $(w/2, h)$ to get $V_{fs} \in \mathbb{R}^{\alpha \times r}$, where the LiDAR data range is set as $r=\sqrt{(w/2)^2 + h^2}$ and angle bins $\alpha$ are set to 180. 

To complete this, for row $j$, we traverse each column $i$ from $0$ to $r$. We first set $V_{fs}(0, j)$ as an invalid area. Then, if $V_{fs}(i, j)$ is an empty area, we complete it as $V_{fs}(i, j) = V_{fs}(i-1, j)$. In other words, if the current pixel in the freespace map is empty and the former column pixel is valid, we replace current pixel with valid, the same as when the former pixel is invalid. After completing each row, we convert the completed freespace map $V_{fs}$ back to the Cartesian coordinates system to obtain the dense freespace map $M_{fs}$. 
$M_{fs}$ is attached to each scene as an additional resource.
Figure \ref{fig_freespace_gen} illustrates the freespace generation process.

\subsection{Recomposition process}
    
\subsubsection{Object-Scene Recomposition}\label{sec:Object-Scene Recomposition}
In this process, we randomly insert objects into freespace of the scene to generate a recomposed scene image.
We first sample a scene from the scene database. Then, we randomly sample $N_{re}$ valid areas according to the scene freespace map. For each valid area, we first get its center bird's-eye-view position $(x_c, z_c)$, and then add a random offset on it to get a sampled position:

\begin{equation}
\begin{aligned}
    (x_s, z_s) &= (x_c, z_c) + (x', y'), \\
    x', y' &\sim \mathcal{U}(-\frac{r_{fs}}{2}, \frac{r_{fs}}{2}).
\end{aligned}
\label{eq:rs_bev_pos}
\end{equation}

For each sampled position $(x_s, z_s)$, we randomly sample an object with a raw bird's-eye-view position $(x_r, z_r)$ from the object database. $(x_r, z_r)$ satisfies the following conditions:

\begin{equation}
\begin{array}{c}
    x_s * x_r > 0, \\
    z_s > z_r * (1 - d_r).
\end{array}
\label{eq:obj_seleted_cond}
\end{equation}

The first $x$-axis condition aims to restrict $x_r$ to share the same sign as $x_s$ because the 3D object model is partly visible. The second $z$-axis condition aims to constrain distant objects from being sampled too close consider to the sparsity of the 3D object points model. 

For each sampled position $(x_s, z_s)$, we refresh the position label of the sampled object from the raw label $(x_r, y_r, z_r)$ to $(x_s, G(x_s, z_s), z_s)$ to keep objects attached to the ground, where $G$ is the ground plane equation of the scene. We keep the 3D size label and the object 3D box yaw angle label invariant. We update the position of each point in the model with the offset:
\begin{equation}
\begin{array}{c}
    \textbf{P}_{offset}=[x_s, f_{ground}(x_s, z_s), z_s]^T - [x_r, y_r, z_r]^T \\
    \textbf{P}_{og} = \textbf{P}_{og} + \textbf{P}_{offset}.
\end{array}
\label{eq:obj_model_resample}
\end{equation}

After collecting all sampled objects, we filter out the collided and occluded objects. We attempt to insert each object from near to far. If the inserted object collides with existing objects, we discard it.

We then project $\textbf{P}_{og}$ and $\textbf{P}_{oc}$ back to the 2D image to obtain the foreground depth map and image patch for each object. We will fill the hole with the nearest projected pixels.

After inpainting the object, we apply an object occlusion check. If one attempt causes any other objects occluded by a large area, the object will be discarded. Specifically, $F_{o}$ denotes the foreground pixel set of an object, and $F_{s}$ denotes the set of pixels occluded by the background of the scene and other objects. We have $F_{s} \subseteq F_{o}$. Given an occlusion threshold $\tau_{o}$, the object will be considered to be largely occluded if $area(F_{s})/area(F_{o}) \textgreater \tau_{o}$. If the inserted object is valid, we merge the foreground depth and image patch with the scene depth and image to get recomposed depth $D_{rs}$ and image $I_{rs}$ according to the pixel-wise depth buffer.

\begin{figure}[]
\centering
\includegraphics[width=1.0\columnwidth]{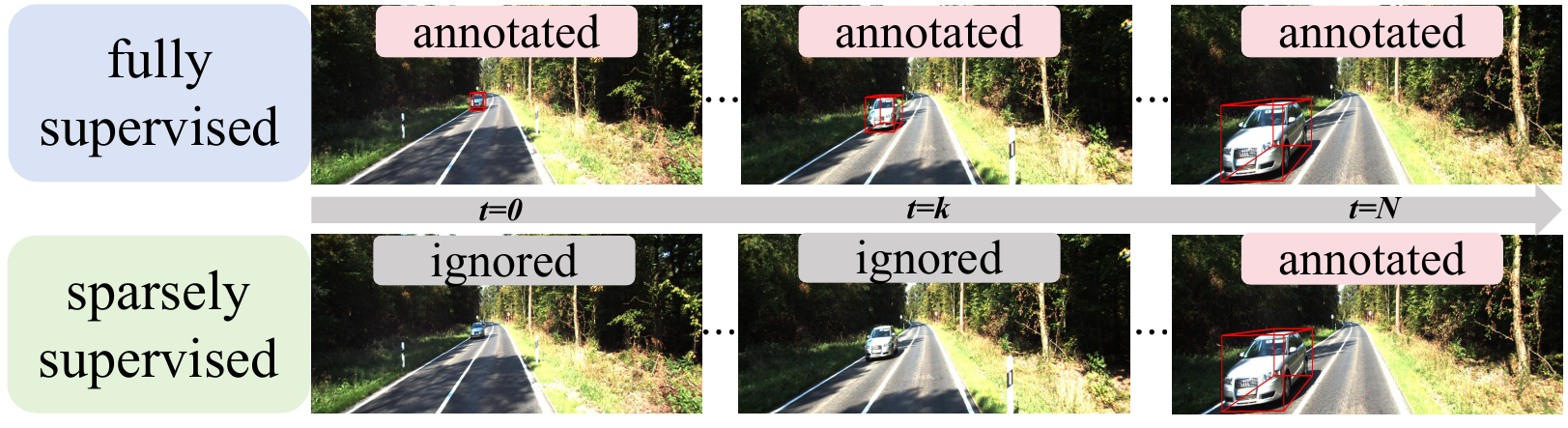} 
\caption{Illustration of the sparsely-supervised setting. 
Based on our scheme, only the objects closest to the camera will be labeled to reduce annotation costs.}
\label{fig_sparsely_sup}
\end{figure}

\subsubsection{Camera pose perturbation}\label{sec:Camera pose perturbation}
The data of the raw scene is entangled with a specific camera pose. We aim to further perturb the camera pose and render the scene from novel views. 

To achieve this, we first decompose the dense depth map $D_{rs}$ and the image $I_{rs}$ from the camera using intrinsic parameters to get scene points $\textbf{P}_{rsg}=T_{2d \xrightarrow{}3d}(D_{rs}, K)$ and corresponding texture $\textbf{P}_{rsc}$ where $\textbf{P}_{rsg} \in \mathbb{R}^{HW \times 3}$, $\textbf{P}_{rsc} \in \mathbb{R}^{HW \times 3}$ and $H, W$ denote the image size. Next, we need to resample a camera pose. We equivalently convert the camera pose perturbation into a coordinate perturbation of the scene point cloud. 

The camera pose has six degrees of freedom, including pitch angle $\theta$, yaw angle $\gamma$, roll angle $\alpha$ for rotation related to $x, y$ and $z$ axis, and translation $\epsilon_x, \epsilon_y, \epsilon_z$ related to $x, y,$ and $z$ axis. Since the mounting position of the camera during training and inference is unchanged, we do not apply the translation perturbation $\epsilon_y$ along the $y$ axis. Since the yaw angle $\gamma$ and the translation $\epsilon_x, \epsilon_z$ along the $x$ and $z$ axes are all along the horizontal direction, we only perturb the $z$ direction.

Specifically, we sample the pitch and roll angles $\theta$ and $\alpha$ for rotation related to the $x$ and $z$ axes as well as the translation perturbation $\epsilon_z$ along the $z$-axis, obtaining a perturbation matrix $\mathbf{R}$ and $\mathbf{t}$ as described in equation (\ref{eq:(5)}) and (\ref{eq:(6)}). The point cloud of the scene, $\mathbf{P}_{rsg}$ and the 3D position labels of the $k$ objects $\mathbf{O} \in \mathbb{R}^{k \times 3} $ in the scene are updated using the new pose as equations (\ref{eq:(7)}) and (\ref{eq:(8)}). 

\begin{gather} 
\mathbf{R}=\begin{bmatrix}
1 & 0 & 0 \\
0 & cos\theta & -sin\theta \\
0 & sin\theta & cos\theta
\end{bmatrix}
\begin{bmatrix}
cos\alpha & -sin\alpha & 0 \\
sin\alpha & cos\alpha & 0 \\
0 & 0 & 1
\end{bmatrix},
\label{eq:(5)} \\
\mathbf{t}=\begin{bmatrix}
0 & 0 & \epsilon_z \\
\end{bmatrix}^T,
\label{eq:(6)} \\
\mathbf{P}_{rsg}^{'}=\mathbf{R} \mathbf{P}_{rsg} + \mathbf{t},
\label{eq:(7)} \\
\mathbf{O}^{'}=\mathbf{R} \mathbf{O} + \mathbf{t}.
\label{eq:(8)}
\end{gather}

Finally, we render the textured scene point cloud to generate a new image and depth map $\{I_t,D_t\}=T_{3d \xrightarrow{}2d}( \{\textbf{P}^{'}_{rsg}, \textbf{P}_{rsc}\}, K)$ for training. 
This process will result in hole pixels in the image after camera zoom-in and hole areas in the ground after camera zoom-out. To handle these, we use max-pooling to fill the hole pixels and use the nearest non-zero ground pixels to fill the hole areas.

\IJCVRevised{We explain the detailed texture points rendering and filling method in algorithm \ref{alg:recom_render}. We consider the difference in the number of hole pixels for objects during object scene recomposition and for the scene after camera perturbation. The former has fewer hole pixels, while the latter has more. We customize their hole filling processes accordingly.}

\begin{algorithm}[!t]
\caption{\IJCVRevised{Texture Points Rendering and Filling}}
\label{alg:recom_render}
\algorithmicrequire{
\,\,  points positions: $\textbf{P}_{g}$; points color: $\textbf{P}_{c}$; intrinsic matrix: $K$, \newline
\textcolor{white}{\textbf{Input:\quad}}image: $I \in \mathbb{R}^{H\times W\times 3}$, depth: $D \in \mathbb{R}^{H\times W}$ 
\newline  
}
\algorithmicensure{ 
recomposed image $I_r$, recomposed depth $D_r$
}
\begin{algorithmic}[1]
\State calculate points image coordinate  $\textbf{u},\textbf{v},\textbf{z} \gets$ $T_{3d \xrightarrow{}2d}(\textbf{P}_{g}, K)$ 
\State initialize : $I'=\textbf{0} \in \mathbb{R}^{H\times W\times 3}, ,M'=\textbf{0} \in \mathbb{R}^{H\times W}$;
\State initialize : $I'[\textbf{v},\textbf{u}]=\textbf{P}_{c}$, $M'[\textbf{v},\textbf{u}]=1$;
\Procedure{object scene recomposition filling}{} 
\State initialize :$D'=\textbf{inf} \in \mathbb{R}^{H\times W}, D'[\textbf{v},\textbf{u}]=\textbf{z}$;
\State fill: $M_{fg}\gets \texttt{MorphologicalHoleFilling}(M')$;
\State extract hole pixels: $\textbf{u}_h, \textbf{u}_h \gets \texttt{Where}(M_{fg}\oplus M'=1)$;
\State  $\textbf{u}_f, \textbf{u}_f \gets \texttt{FindNearsetPoint}((\textbf{u}_h, \textbf{u}_h),(\textbf{u},\textbf{v}))$;
\State fill: $I'[\textbf{u}_h, \textbf{u}_h]=I'[\textbf{u}_f, \textbf{u}_f], D'[\textbf{u}_h, \textbf{u}_h]=D'[\textbf{u}_f, \textbf{u}_f]$;
\State $I_r,D_r \gets \texttt{ZBufferMerging}(D',D,I',I) $
\EndProcedure

\Procedure{camera pose perturbation filling}{} 
\State initialize :$D'=\textbf{0} \in \mathbb{R}^{H\times W}, D'[\textbf{v},\textbf{u}]=\textbf{z}$;
\State $I_r,D_r \gets \texttt{GPUMaxPooling}(I', D')$;
\State $I_r,D_r \gets \texttt{GPUGaussianSmoothing}(I_r, D_r)$;
\State $I_r[M']=I'[M'], D_r[M']=D'[M']$
\EndProcedure

\State \textbf{return} $I_r, D_r$ 

\end{algorithmic}
\end{algorithm}

\subsubsection{Mix sampling of Raw/Empty Scenes}\label{sec:Mix sampling of Raw/Empty Scenes}
On one hand, using only raw scenes for recomposition is inflexible due to the occupancy of existing objects. On the other hand, using only empty scenes will induce a domain gap between real and synthetic data. To make a trade-off, we employ a mixed sampling strategy of raw scenes and empty scenes during training. Specifically, for a batch size $bs$, we randomly sample $(1-r_{empty}) \times bs$ raw scenes and $r_{empty} \times bs$ empty scenes to perform our recomposition process and obtain training data. 

\subsection{Applications}

Our object-scene-camera decomposition and recomposition data manipulation scheme can serve as a plug-and-play component to boost the performance of M3OD models in flexible annotation settings.

\subsubsection{Fully-Supervised Setting}
\label{sec:fully-supervised-setting}

In the fully-supervised setting, all frames in all collected clips will be fully annotated. These frames are treated as raw scenes. All scenes will be used to generate empty scenes. Raw scenes and empty scenes are both add to the scene database. Objects in all these frames are collected into the object database. Notably, we only select high-quality objects with depths less than 50 meters, slight occlusion, and no truncation to construct the object database.

\subsubsection{Sparsely-Supervised Setting}
Fully annotating all frames is costly and limits real-world applications. We propose a sparsely-supervised scheme, where only the closest object with the richest semantic information is labeled for each instance in each clip as shown in Figure \ref{fig_sparsely_sup}. Using a 2D tracking method ~\citep{maggiolino2023deep}, we extract instances and select only the closest object to label, identified by LiDAR points in the foreground. These sparse objects, which make up about 5\% of all annotations, are used to build the object database denoted $DB_{obj}^{sparse}$. All frames are used to construct the empty scene database, denoted $DB_{scene}^{empty}$.

To handle the domain gap of real and synthetic data, we can randomly select a few clips to fully annotate. 
The amount of full annotations can also be flexibly controlled for different applications in the real world. 
More specifically, all labeled objects in our scheme account for $k \%$ of all labeled objects in the KITTI train split, with $k$ being flexible to control annotation costs.
The scenes in the selected clips serve as raw scenes to construct the raw scene database $DB_{scene}^{raw}$. High-quality objects in these selected scenes are used to construct the object database $DB_{obj}^{raw}$.

Because the amount of empty scenes and raw scenes is unbalanced in the sparsely-supervised setting, we train the model for two stages. 
In the first stage, we use $DB_{obj}^{sparse}$ and $DB_{scene}^{empty}$. During each epoch, all scenes in $DB_{scene}^{empty}$ are traversed to recompose with $DB_{obj}^{sparse}$ and camera poses, generating new training images for model pretraining. 
In the second stage, we merge $DB_{obj}^{sparse}$ and $DB_{obj}^{raw}$ to obtain $DB_{obj}$. During each epoch, we randomly select half the scenes from $DB_{scene}^{raw}$ and the same number of scenes in $DB_{scene}^{empty}$ to recompose with $DB_{obj}$  and camera poses, generating new training images for model finetuning.
\begin{table*}[]
\caption{Comparison with four base models on the car category in the KITTI official test set and validation split.}
\centering
\begin{tabular}{@{}c|cccccc|cccccc@{}}
\toprule
\multirow{2}{*}{\textbf{Method}} & \multicolumn{3}{c}{\textbf{$AP_{3D}^{test}$}}  & \multicolumn{3}{c|}{\textbf{$AP_{BEV}^{test}$}} & \multicolumn{3}{c}{\textbf{$AP_{3D}^{val}$}}   & \multicolumn{3}{c}{\textbf{$AP_{BEV}^{val}$}}   \\
                                 & easy           & mod            & hard           & easy            & mod            & hard           & easy           & mod            & hard           & easy            & mod            & hard           \\ \midrule
MonoDLE                          & 17.23          & 12.23          & 10.29          & 24.79           & 18.89          & 16.00          & 19.48          & 15.21          & 13.57          & 26.03           & 21.10          & 18.59          \\
+ours                            & 25.55          & 18.04          & 15.45          & 34.51           & 24.82          & 20.86          & 26.57          & 20.42          & 17.34          & 36.37           & 26.74          & 23.07          \\
\textbf{improvement}             & \textbf{+8.32} & \textbf{+5.81} & \textbf{+5.16} & \textbf{+9.72}  & \textbf{+5.93} & \textbf{+4.86} & \textbf{+7.09} & \textbf{+5.21} & \textbf{+3.77} & \textbf{+10.34} & \textbf{+5.64} & \textbf{+4.48} \\ \midrule
GUPNet                           & 22.26          & 15.02          & 13.12          & 30.29           & 21.19          & 18.20          & 21.95          & 16.94          & 14.43          & 29.57           & 23.49          & 20.40          \\
+ours                            & 28.60          & 19.35          & 16.35          & 38.18           & 26.51          & 23.01          & 26.91          & 20.45          & 17.45          & 35.17           & 26.22          & 22.84          \\
\textbf{improvement}             & \textbf{+6.34} & \textbf{+4.33} & \textbf{+3.23} & \textbf{+7.89}  & \textbf{+5.32} & \textbf{+4.81} & \textbf{+4.96} & \textbf{+3.51} & \textbf{+3.02} & \textbf{+5.60}   & \textbf{+2.73} & \textbf{+2.44} \\ \midrule
DID-M3D                          & 24.40          & 16.29          & 13.75          & 32.95           & 22.76          & 19.83          & 24.60          & 17.30          & 14.57          & 31.26           & 22.90          & 20.46          \\
+ours                            & 30.69          & 20.45          & 17.25          & 39.68           & 26.50          & 22.69          & 30.32          & 21.65          & 18.28          & 39.03           & 27.79          & 23.93          \\
\textbf{improvement}             & \textbf{+6.29} & \textbf{+4.16} & \textbf{+3.50}  & \textbf{+6.73}  & \textbf{+3.74} & \textbf{+2.86} & \textbf{+5.72} & \textbf{+4.35} & \textbf{+3.71} & \textbf{+7.77}  & \textbf{+4.89} & \textbf{+3.47} \\ \midrule
MonoDETR                         & 25.00          & 16.47          & 13.58          & 33.60           & 22.11          & 18.60          & 28.84          & 20.61          & 16.38          & 37.86           & 26.95          & 22.80          \\
+ours                            & 30.25          & 20.86          & 17.68          & 38.94           & 26.75          & 23.13          & 31.58          & 23.43          & 20.01          & 42.35           & 30.94          & 26.05          \\
\textbf{improvement}             & \textbf{+5.25} & \textbf{+4.39} & \textbf{+4.10} & \textbf{+5.34}  & \textbf{+4.64} & \textbf{+4.53} & \textbf{+2.74} & \textbf{+2.82} & \textbf{+3.63} & \textbf{+4.49}  & \textbf{+3.99} & \textbf{+3.25} \\ \midrule
\IJCVRevised{MonoDGP}                         & 26.35          & 18.72          & 15.97          & 35.24           & 25.23          & 22.02          & 30.76          & 22.34          & 19.02          & 39.40           & 28.20          & 24.42          \\
+ours                            & 30.25          & 21.10          & 17.85          & 37.95           & 26.82          & 23.11          & 32.02          & 23.90          & 20.49          & 40.90           & 30.80          & 26.95          \\
\textbf{improvement}             & \textbf{+3.90} & \textbf{+2.38} & \textbf{+1.88} & \textbf{+2.71}  & \textbf{+1.59} & \textbf{+1.09} & \textbf{+1.26} & \textbf{+1.56} & \textbf{+1.47} & \textbf{+1.50}  & \textbf{+2.60} & \textbf{+2.53} \\ \bottomrule
\end{tabular}
\label{table_baseline_cmp_pro}
\end{table*}

\setul{0.2ex}{}

\begin{table*}[]
\caption{Comparison with state-of-the-art methods on the car category in the KITTI official test set. Following the KITTI protocol, methods are ranked based on their performance under the moderate difficulty setting. The best results are in \textbf{bold}, and the best results of the baselines are \ul{underlined}.}
\centering
\begin{tabular}{@{}
>{\columncolor[HTML]{FFFFFF}}c |
>{\columncolor[HTML]{FFFFFF}}c |
>{\columncolor[HTML]{FFFFFF}}c 
>{\columncolor[HTML]{FFFFFF}}c 
>{\columncolor[HTML]{FFFFFF}}c |
>{\columncolor[HTML]{FFFFFF}}c 
>{\columncolor[HTML]{FFFFFF}}c 
>{\columncolor[HTML]{FFFFFF}}c @{}}
\toprule
\cellcolor[HTML]{FFFFFF}                                  & \cellcolor[HTML]{FFFFFF}                                  & \multicolumn{3}{c|}{\cellcolor[HTML]{FFFFFF}$AP_{3D}$} & \multicolumn{3}{c}{\cellcolor[HTML]{FFFFFF}$AP_{BEV}$} \\
\multirow{-2}{*}{\cellcolor[HTML]{FFFFFF}\textbf{Method}} & \multirow{-2}{*}{\cellcolor[HTML]{FFFFFF}\textbf{Venues}} & easy                 & mod                  & hard                & easy                 & mod                  & hard                \\ \midrule
MonoDLE ~\citep{ma2021delving}              & CVPR2021                                                  & 17.23                & 12.23                & 10.29               & 24.79                & 18.89                & 16.00               \\
CaDDN ~\citep{reading2021categorical}       & CVPR2021                                                  & 19.17                & 13.41                & 11.46               & 27.94                & 18.91                & 17.19               \\
MonoFlex ~\citep{zhang2021objects}          & CVPR2021                                                  & 19.94                & 13.89                & 12.07               & 28.23                & 19.75                & 16.89               \\
GUPNet ~\citep{lu2021geometry}              & ICCV2021                                                  & 22.26                & 15.02                & 13.12               & 30.29                & 21.19                & 18.20               \\
MonoDistill ~\citep{chong2022monodistill}   & ICLR2022                                                  & 22.97                & 16.03                & 13.60               & 31.87                & 22.59                & 19.72               \\
DID-M3D ~\citep{peng2022did}                & ECCV2022                                                  & 24.40                & 16.29                & 13.75               & 32.95                & 22.76                & 19.83               \\
Monocon ~\citep{liu2022learning}            & AAAI2022                                                  & 22.50                & 16.46                & 13.95               & 31.12                & 22.10                & 19.00               \\
CMKD ~\citep{hong2022cross}                 & ECCV2022                                                  & 25.09                & 16.99                & 15.30               & 33.69                & 23.10                & 20.67               \\
MonoDDE ~\citep{li2022diversity}            & CVPR2022                                                  & 24.93                & 17.14                & 15.10               & 33.58                & 23.46                & 20.37               \\
MonoDETR ~\citep{zhang2023monodetr}         & ICCV2023                                                  & 24.52                & 16.26                & 13.93               & 32.20                & 21.45                & 18.68               \\
MonoNeRD ~\citep{zhang2023monodetr}         & ICCV2023                                                  & 22.75                & 17.13                & 15.63               & 31.13                & 23.46                & 20.97               \\
NeurOCS ~\citep{min2023neurocs}             & CVPR2023                                                  & \ul{29.89}          & 18.94                & 15.90               & \ul{37.27}          & 24.49                & 20.89               \\
MonoCD ~\citep{yan2024monocd}               & CVPR2024                                                  & 25.53                & 16.59                & 14.53               & 33.41                & 22.81                & 19.57               \\
OccupancyM3D ~\citep{peng2024learning}      & CVPR2024                                                  & 25.55                & 17.02                & 14.79               & 35.38                & 24.18                & 21.37               \\
MonoSTL ~\citep{ding2024selective}          & TCSVT2024                                                 & 24.54                & 17.14                & 14.59               & 32.19                & 22.42                & 19.48               \\
MonoLSS ~\citep{li2024monolss}              & 3DV2024                                                   & 26.11                & \ul{19.15}          & \ul{16.94}         & 34.89                & \ul{25.95}          & \ul{22.59}         \\
MonoDGP ~\citep{pu2025monodgp}              & CVPR2025                                                   & 26.35                & 18.72          & 15.97         & 35.24                & 25.23          & 22.02         \\ \midrule
MonoDLE+ours                                              & \cellcolor[HTML]{FFFFFF}                                  & 25.55                & 18.04                & 15.45               & 34.51                & 24.82                & 20.86               \\
GUPNet+ours                                               & \cellcolor[HTML]{FFFFFF}                                  & 28.60                & 19.35                & 16.35               & 38.18                & 26.51                & 23.01               \\
DID-M3D+ours                                              & \cellcolor[HTML]{FFFFFF}                                  & \textbf{30.69}       & 20.45                & 17.25               & \textbf{39.68}       & 26.50                & 22.69               \\
MonoDETR+ours                                             & \cellcolor[HTML]{FFFFFF}                                  & 30.25                & 20.86       & 17.68      & 38.94                & 26.75       & \textbf{23.13}      \\
\textbf{\IJCVRevised{MonoDGP+ours}}                                             & \cellcolor[HTML]{FFFFFF}                                  & 30.25                & \textbf{21.10}       & \textbf{17.85}      & 37.95                & \textbf{26.82}       & 23.11      \\ \cmidrule(r){1-1} \cmidrule(l){3-8} 
vs. SOTAs                                                 & \multirow{-5}{*}{\cellcolor[HTML]{FFFFFF}-}               & +0.80                 & +1.95                & +0.91               & +2.41                & +0.87                & +0.54               \\ \bottomrule
\end{tabular}
\label{table_kitti_testset_pro}
\end{table*}

\section{Experiments}\label{sec:exp}

\subsection{Datasets and Metrics}
To demonstrate the effectiveness of our data manipulation scheme, we evaluate it on both the widely-used KITTI dataset ~\citep{geiger2012we} and the more complicated Waymo dataset ~\citep{sun2020scalability}.

The KITTI 3D object detection dataset~\citep{geiger2012we} provides 141 clips and 7481 images for training and 7518 images for testing. Following common practice ~\citep{ma2021delving}, we split the training set into a train split and a validation split. The train split contains 96 clips and 3712 images, while the validation split contains 45 clips and 3769 images. The image resolution is set to 384$\times$1280. Following previous methods, we use Average Precision $AP_{3D}^{R40}$ and $AP_{BEV}^{R40}$ to evaluate the validation split and the test set.

The Waymo Dataset~\citep{sun2020scalability} is a more recent large-scale dataset that consists of 798 training clips and 202 validation clips. The dataset provides objects labels in the full 360° field of view with a multi-camera rig. \IJCVRevised{Waymo has 5 ring cameras covering a 230° field of view to perceive the environment. To evaluate our method under both monocular and multi-camera settings, we separately discuss two configurations of Waymo: Waymo-Mono and Waymo-Ring.}

\IJCVRevised{For Waymo-Mono, } following ~\citep{reading2021categorical}, we use only the front camera and consider only object labels in the front camera's field of view for the task of M3OD. We provide results on the validation clips. Considering the large dataset size and high frame rate \IJCVRevised{(10Hz)}, we sample every 3\textsuperscript{rd} frame from the training clips to form our training set (52,386 images) following ~\citep{reading2021categorical}. The image resolution is downsampled to 640$\times$960 to meet GPU memory. We adopt the official evaluation to calculate the Average Precision $AP$ and Average Precision weighted by Heading $APH$. The evaluation is divided by difficulty settings (LEVEL\_1, LEVEL\_2), which are determined by the number of LiDAR points inside the 3D box. We evaluate using IoU criteria of 0.7 and 0.5.

\IJCVRevised{For Waymo-Ring, we use all 5 ring cameras and extract object labels within the filed of view. We sample every 4\textsuperscript{th} frame for the training clips to form the training set (39872$\times$ 5 images). The image resolution is downsampled to 352$\times$800. we follow the popular multi-camera evaluation protocol used by Nuscenes~\citep{caesar2020nuscenes}. We use the metric NDS$^*$, which aggregates mean Average Precision (mAP), mean Average Translation Error (mATE), mean Average Scale Error (mASE), and mean Average Orientation Error (mAOE):}
\begin{equation}
    \IJCVRevised{\text{NDS}^*=\frac{1}{6}[3\text{mAP}+\sum_{\text{mTP}\in \mathbb{TP}}(1-\text{min}(1, \text{mTP}))].}
    \label{eq:nds*}
\end{equation}

\subsection{Implementation Details}
\label{sec:exp_implementation_details}
In the experiments, we empirically set the parameters. During object edge rectification, $n_{nb}$ in Equation (1) is set to 5. 
For scene freespace map construction in Section \ref{sec:Scene freespace generation}, $r_{fs}$ is set to 0.5m and the size $(h,w)$ of the freespace map is set to (140, 160).
During object-scene recomposition in Section \ref{sec:Object-Scene Recomposition}, $N_{re}$ is randomly sampled from 0 to 10 for a raw scene and 5 to 15 for an empty scene, and $\tau_{o}$ is randomly sampled from $\{0.1, 0.3, 0.5, 0.7\}$. During the perturbation of camera poses in Section \ref{sec:Camera pose perturbation}, we randomly set $\theta$ and $\alpha$ within $[-2^{\circ}, 2^{\circ}]$, and $\epsilon_{z}$ within [-2m, 2m]. During training in Section \ref{sec:Mix sampling of Raw/Empty Scenes}, we set $r_{empty}$ to 0.5. \IJCVRevised{Moreover, for Waymo-Ring, $r_{fs}$ is set to 0.4m and the size $(h,w)$ of the freespace map is set to (256, 256). $N_{re}$ will be randomly sampled to control the range of training objects from 10 to 50 for a scene.} 

\IJCVRevised{To filter out the high-quality objects as mentioned in \ref{sec:fully-supervised-setting}, we use the occlusion and truncation labels in KITTI and determine the labels in Waymo using simple rules. Specifically, for KITTI, we filter out objects with truncation greater than 0.5 and occlusion greater than 2. For Waymo, to determine the truncation, we first project the 3D box corners into image space and calculate the 2D box area $A_{box}$. We then calculate the area $A_{box}'$ inside the image. Next, we filter out objects with large truncation that satisfy $\frac{A_{box}'}{A_{box}}<0.9$. Moreover, let the segmentation area of an object be denoted as $A_{seg}$, we filter out objects with large occlusion that satisfy $\frac{A_{seg}}{A_{box}'}<0.5$ for Vehicles and $\frac{A_{seg}}{A_{box}'}<0.3$ for Pedestrians and Cyclists. For Waymo-Mono and Waymo-Ring, we use the sparsely annotated objects database for all settings, considering the large amount of data.}

Our model is trained using a one-cycle learning rate scheduler and the AdamW optimizer with a single RTX 4090 GPU. We train the model for 200 epochs on the KITTI dataset and 30 epochs on the Waymo dataset. The batch size is set to 16. Notably, all parameters can be further optimized for different base models to boost the performance. \IJCVRevised{For Waymo-Ring, we train the model for 24 epochs on four RTX 4090 GPUs. The total batch size is set to 32.}

\begin{table*}[]
\caption{Comparison of fully-supervised and sparsely-supervised results on the KITTI validation set. Our scheme with fewer annotations (especially 10\% for the first three base models) achieves comparable performance to the fully-supervised training of the base models. When both use the same sparse annotations, our scheme significantly outperforms the base models.}
\centering
\begin{tabular}{@{}c|c|ccc|ccc@{}}
\toprule
\multirow{2}{*}{\textbf{Method}}   & \multirow{2}{*}{\textbf{Annotation}} & \multicolumn{3}{c|}{$AP_{3D}$}                                      & \multicolumn{3}{c}{$AP_{BEV}$}                                      \\
                          &                             & easy                  & mod                   & hard                  & easy                  & mod                   & hard                  \\ \midrule
\multirow{3}{*}{MonoDLE}  & Full                        & 19.48                 & 15.21                 & 13.57                 & 26.03                 & 21.10                 & 18.59                 \\
                          & 10\%                        & 6.67(-12.81)          & 4.67(-10.54)          & 3.61(-9.96)           & 10.10(-15.93)         & 6.91(-14.19)          & 5.54(-13.05)          \\
                          & \textbf{10\% w / ours}      & \textbf{20.73(+1.25)} & \textbf{15.73(+0.52)}  & \textbf{12.82(-0.75)} & \textbf{28.00(+1.97)}  & \textbf{20.69(-0.41)} & \textbf{16.37(-2.22)} \\ \midrule
\multirow{3}{*}{GUPNet}   & Full                        & 21.95                 & 16.94                 & 14.43                 & 29.57                 & 23.49                 & 20.40                 \\
                          & 10\%                        & 10.25(-11.70)          & 6.06(-10.88)          & 4.70(-9.73)           & 15.17(-14.40)          & 8.88(-14.61)          & 7.13(-13.27)          \\
                          & \textbf{10\% w / ours}      & \textbf{22.27(+0.32)} & \textbf{16.18(-0.76)} & \textbf{13.18(-1.25)} & \textbf{29.94(+0.37)} & \textbf{21.48(-2.01)} & \textbf{17.03(-3.37)} \\ \midrule
\multirow{3}{*}{DID-M3D}  & Full                        & 24.60                 & 17.30                 & 14.57                 & 31.26                 & 22.90                 & 20.46                 \\
                          & 10\%                        & 13.05(-11.55)         & 8.50(-8.80)            & 6.13(-8.44)           & 18.32(-12.94)         & 11.64(-11.26)         & 8.70(-11.76)          \\
                          & \textbf{10\% w / ours}      & \textbf{24.74(+0.14)} & \textbf{17.22(-0.08)} & \textbf{13.99(-0.58)} & \textbf{32.62(+1.36)} & \textbf{22.60(-0.30)}  & \textbf{18.74(-1.72)} \\ \midrule
\multirow{3}{*}{MonoDETR} & Full                        & 28.84                 & 20.61                 & 16.38                 & 37.86                 & 26.95                 & 22.80                 \\
                          & 50\%                        & 18.09(-10.75)         & 13.39(-7.22)          & 11.53(-4.85)          & 28.01(-9.85)          & 20.70(-6.25)          & 17.41(-5.39)          \\
                          & \textbf{50\% w / ours}      & \textbf{28.23(-0.61)} & \textbf{20.59(-0.02)} & \textbf{17.23(+0.85)} & \textbf{37.42(-0.44)} & \textbf{27.59(+0.64)} & \textbf{23.56(+0.76)} \\ \bottomrule
\end{tabular}
\label{table_10label_4bsl}
\end{table*}

\begin{table*}[t]
\caption{Comparison of base models and our scheme on three categories in the \IJCVRevised{Waymo-Mono} Dataset.}
\centering
\begin{tabular}{@{}
>{\columncolor[HTML]{FFFFFF}}c |
>{\columncolor[HTML]{FFFFFF}}c |
>{\columncolor[HTML]{FFFFFF}}c |
>{\columncolor[HTML]{FFFFFF}}c 
>{\columncolor[HTML]{FFFFFF}}c 
>{\columncolor[HTML]{FFFFFF}}c |
>{\columncolor[HTML]{FFFFFF}}c 
>{\columncolor[HTML]{FFFFFF}}c 
>{\columncolor[HTML]{FFFFFF}}c @{}}
\toprule
\cellcolor[HTML]{FFFFFF}                                        & \cellcolor[HTML]{FFFFFF}                                      & \cellcolor[HTML]{FFFFFF}                                  & \multicolumn{3}{c|}{\cellcolor[HTML]{FFFFFF}$AP_{3D}$}    & \multicolumn{3}{c}{\cellcolor[HTML]{FFFFFF}$APH_{3D}$}    \\
\multirow{-2}{*}{\cellcolor[HTML]{FFFFFF}\textbf{$IOU_{3D}$}} & \multirow{-2}{*}{\cellcolor[HTML]{FFFFFF}\textbf{Difficulty}} & \multirow{-2}{*}{\cellcolor[HTML]{FFFFFF}\textbf{Method}} & Vehicle      & Pedestrian  & Cyclist      & Vehicle      & Pedestrian  & Cyclist      \\ \midrule
\cellcolor[HTML]{FFFFFF}                                        & \cellcolor[HTML]{FFFFFF}                                      & DID-M3D                                                   & 2.64                  & 1.49                 & 1.05                  & 2.63                  & 1.36                 & 1.02                  \\
\cellcolor[HTML]{FFFFFF}                                        & \cellcolor[HTML]{FFFFFF}                                      & \textbf{+ours}                                            & \textbf{3.73(+41\%)}  & \textbf{1.91(+28\%)} & \textbf{2.78(+165\%)} & \textbf{3.71(+41\%)}  & \textbf{1.72(+26\%)} & \textbf{2.70(+165\%)} \\
\cellcolor[HTML]{FFFFFF}                                        & \cellcolor[HTML]{FFFFFF}                                      & MonoDETR                                                  & 1.63                  & 0.98                 & 1.87                  & 1.61                  & 0.86                 & 1.49                  \\
\cellcolor[HTML]{FFFFFF}                                        & \multirow{-4}{*}{\cellcolor[HTML]{FFFFFF}Level\_1}            & \textbf{+ours}                                            & \textbf{2.52(+55\%)}  & \textbf{1.34(+37\%)} & \textbf{2.14(+14\%)}  & \textbf{2.49(+55\%)}  & \textbf{1.2(+40\%)}  & \textbf{2.1(+41\%)}   \\ \cmidrule(l){2-9} 
\cellcolor[HTML]{FFFFFF}                                        & \cellcolor[HTML]{FFFFFF}                                      & DID-M3D                                                   & 2.48                  & 1.36                 & 1.01                  & 2.46                  & 1.24                 & 0.98                  \\
\cellcolor[HTML]{FFFFFF}                                        & \cellcolor[HTML]{FFFFFF}                                      & \textbf{+ours}                                            & \textbf{3.50(+41\%)}  & \textbf{1.74(+27\%)} & \textbf{2.68(+165\%)} & \textbf{3.47(+41\%)}  & \textbf{1.57(+27\%)} & \textbf{2.60(+165\%)} \\
\cellcolor[HTML]{FFFFFF}                                        & \cellcolor[HTML]{FFFFFF}                                      & MonoDETR                                                  & 1.53                  & 0.89                 & 1.80                   & 1.51                  & 0.78                 & 1.43                  \\
\multirow{-8}{*}{\cellcolor[HTML]{FFFFFF}0.7}                   & \multirow{-4}{*}{\cellcolor[HTML]{FFFFFF}Level\_2}            & \textbf{+ours}                                            & \textbf{2.37(+55\%)}  & \textbf{1.22(+37\%)} & \textbf{2.06(+14\%)}  & \textbf{2.34(+55\%)}  & \textbf{1.09(+40\%)} & \textbf{2.02(+41\%)}  \\ \midrule
\cellcolor[HTML]{FFFFFF}                                        & \cellcolor[HTML]{FFFFFF}                                      & DID-M3D                                                   & 11.83                 & 7.17                 & 5.42                  & 11.74                 & 6.41                 & 5.26                  \\
\cellcolor[HTML]{FFFFFF}                                        & \cellcolor[HTML]{FFFFFF}                                      & \textbf{+ours}                                            & \textbf{14.42(+22\%)} & \textbf{8.13(+13\%)} & \textbf{8.30(+53\%)}  & \textbf{14.30(+22\%)} & \textbf{7.20(+12\%)} & \textbf{7.99(+52\%)}  \\
\cellcolor[HTML]{FFFFFF}                                        & \cellcolor[HTML]{FFFFFF}                                      & MonoDETR                                                  & 8.03                  & 4.77                 & 6.39                  & 7.94                  & 4.12                 & 5.58                  \\
\cellcolor[HTML]{FFFFFF}                                        & \multirow{-4}{*}{\cellcolor[HTML]{FFFFFF}Level\_1}            & \textbf{+ours}                                            & \textbf{11.08(+38\%)} & \textbf{6.18(+30\%)} & \textbf{6.47(+1\%)}   & \textbf{10.95(+38\%)} & \textbf{5.45(+32\%)} & \textbf{6.30(+13\%)}   \\ \cmidrule(l){2-9} 
\cellcolor[HTML]{FFFFFF}                                        & \cellcolor[HTML]{FFFFFF}                                      & DID-M3D                                                   & 11.09                 & 6.53                 & 5.22                  & 11.01                 & 5.83                 & 5.06                  \\
\cellcolor[HTML]{FFFFFF}                                        & \cellcolor[HTML]{FFFFFF}                                      & \textbf{+ours}                                            & \textbf{13.52(+22\%)} & \textbf{7.40(+13\%)} & \textbf{8.00(+53\%)}  & \textbf{13.41(+22\%)} & \textbf{6.55(+12\%)} & \textbf{7.70(+52\%)}  \\
\cellcolor[HTML]{FFFFFF}                                        & \cellcolor[HTML]{FFFFFF}                                      & MonoDETR                                                  & 7.54                  & 4.34                 & 6.16                  & 7.45                  & 3.75                 & 5.37                  \\
\multirow{-8}{*}{\cellcolor[HTML]{FFFFFF}0.5}                   & \multirow{-4}{*}{\cellcolor[HTML]{FFFFFF}Level\_2}            & \textbf{+ours}                                            & \textbf{10.39(+38\%)} & \textbf{5.62(+29\%)} & \textbf{6.23(+1\%)}   & \textbf{10.27(+38\%)} & \textbf{4.96(+32\%)} & \textbf{6.07(+13\%)}  \\ \bottomrule
\end{tabular}
\label{table_waymo_did_detr_3cls_res}
\end{table*}

\begin{table*}[t]
\caption{Comparison of fully supervised and sparsely supervised (10\% annotation ratio) settings on \IJCVRevised{Waymo-Mono} dataset (vehicle class). Our scheme with 10\% annotation achieves comparable performance to the fully-supervised setting of the base models.}
\centering
\begin{tabular}{@{}ccc|ccc|ccc@{}}
\toprule
\multicolumn{3}{c|}{\textbf{Method}}                                              & \multicolumn{3}{c|}{DID-M3D}                & \multicolumn{3}{c}{MonoDETR}               \\ \midrule
\multicolumn{3}{c|}{\textbf{Annotation}}                                          & Full  & 10\%        & \textbf{10\% w/ours}  & Full & 10\%        & \textbf{10\% w /ours} \\ \midrule
\multirow{4}{*}{\textbf{$AP_{3D}$}}  & \multirow{2}{*}{$IOU_{3D}@0.7$} & Level\_1 & 2.64  & 0.71(-1.93) & \textbf{2.69(+0.05)}  & 1.63 & 0.77(-0.86) & \textbf{1.53(-0.10)}  \\
                                     &                                 & Level\_2 & 2.48  & 0.67(-1.81) & \textbf{2.52(+0.04)}  & 1.53 & 0.72(-0.81) & \textbf{1.44(-0.09)}  \\
                                     & \multirow{2}{*}{$IOU_{3D}@0.5$} & Level\_1 & 11.83 & 4.37(-7.46) & \textbf{11.24(-0.59)} & 8.03 & 4.51(-3.52) & \textbf{7.47(-0.56)}  \\
                                     &                                 & Level\_2 & 11.09 & 4.09(-7.00) & \textbf{10.54(-0.55)} & 7.54 & 4.23(-3.31) & \textbf{7.01(-0.53)}  \\ \midrule
\multirow{4}{*}{\textbf{$APH_{3D}$}} & \multirow{2}{*}{$IOU_{3D}@0.7$} & Level\_1 & 2.63  & 0.70(-1.93) & \textbf{2.66(+0.03)}  & 1.61 & 0.76(-0.85) & \textbf{1.51(-0.10)}  \\
                                     &                                 & Level\_2 & 2.46  & 0.66(-1.80) & \textbf{2.50(+0.04)}  & 1.51 & 0.71(-0.80) & \textbf{1.42(-0.09)}  \\
                                     & \multirow{2}{*}{$IOU_{3D}@0.5$} & Level\_1 & 11.74 & 4.29(-7.45) & \textbf{11.13(-0.61)} & 7.94 & 4.42(-3.52) & \textbf{7.35(-0.59)}  \\
                                     &                                 & Level\_2 & 11.01 & 4.02(-6.99) & \textbf{10.44(-0.57)} & 7.45 & 4.14(-3.31) & \textbf{6.90(-0.55)}   \\ \bottomrule
\end{tabular}
\label{table_waymo_10_label_did_detr_car}
\end{table*}

\subsection{Main Results on KITTI}

Our data manipulation scheme can serve as a plug-and-play component to boost M3OD models. 
We widely use four recent representative models with released codes, 
including MonoDLE ~\citep{ma2021delving}, GUPNet ~\citep{lu2021geometry}, DID-M3D ~\citep{peng2022did}, and MonoDETR ~\citep{zhang2023monodetr}.

\begin{table*}[t]
\caption{\IJCVRevised{Comparison of both fully-supervised and sparsely supervised performance at different distances. The result is reported on DID-M3D with an IoU threshold of 0.5.}}
\centering
\begin{tabular}{@{}c|c|c|ccc|ccc@{}}
\toprule
\multirow{2}{*}{\textbf{Setting}}       & \multirow{2}{*}{\textbf{Difficulty}} & \multirow{2}{*}{\textbf{Method}} & \multicolumn{3}{c|}{$AP_{3D}$}                      & \multicolumn{3}{c}{$APH_{3D}$}                      \\
                               &                             &                         & 0-30m          & 30-50m        & 50-$\infty$ m         & 0-30m          & 30-50m        & 50-$\infty$ m         \\ \midrule
\multirow{4}{*}{fully-supervised}    & \multirow{2}{*}{Level\_1}   & DID-M3D                 & 27.55          & 5.47          & 0.89          & 27.35          & 5.42          & 0.88          \\
                               &                             & \textbf{ours}           & \textbf{33.68} & \textbf{6.53} & \textbf{0.75} & \textbf{33.41} & \textbf{6.47} & \textbf{0.74} \\ \cmidrule(l){2-9} 
                               & \multirow{2}{*}{Level\_2}   & DID-M3D                 & 27.45          & 5.27          & 0.78          & 27.25          & 5.22          & 0.76          \\
                               &                             & \textbf{ours}           & \textbf{33.55} & \textbf{6.28} & \textbf{0.65} & \textbf{33.29} & \textbf{6.23} & \textbf{0.64} \\ \midrule
\multirow{4}{*}{sparsely-supervised (10\%)} & \multirow{2}{*}{Level\_1}   & DID-M3D                 & 11.84          & 1.41          & 0.18          & 11.64          & 1.38          & 0.17          \\
                               &                             & \textbf{ours}           & \textbf{27.20} & \textbf{4.81} & \textbf{0.42} & \textbf{26.95} & \textbf{4.78} & \textbf{0.42} \\ \cmidrule(l){2-9} 
                               & \multirow{2}{*}{Level\_2}   & DID-M3D                 & 11.80          & 1.36          & 0.16          & 11.60          & 1.33          & 0.15          \\
                               &                             & \textbf{ours}           & \textbf{27.10} & \textbf{4.64} & \textbf{0.37} & \textbf{26.85} & \textbf{4.61} & \textbf{0.36} \\ \bottomrule
\end{tabular}
\label{table_waymo_did_distance}
\end{table*}

\begin{table*}[t]
\caption{\IJCVRevised{Comparison of PETR ~\citep{liu2022petr} and our scheme on Waymo-Ring Dataset.}}
\centering
\begin{tabular}{@{}cc|ccccc|ccccc@{}}
\toprule
\multicolumn{2}{c|}{\textbf{Supervision}}                                  & \multicolumn{5}{c|}{Full}                                                                  & \multicolumn{5}{c}{Sparse(10\%)}                                                           \\ \midrule
\multicolumn{1}{c|}{\textbf{Category}}                    & \textbf{Method}         & mAP$\uparrow$            & mATE$\downarrow$           & mASE$\downarrow$           & mAOE$\downarrow$           & NDS*$\uparrow$                   & mAP$\uparrow$            & mATE$\downarrow$           & mASE$\downarrow$           & mAOE$\downarrow$           & NDS*$\uparrow$                   \\ \midrule
\multicolumn{1}{c|}{\multirow{2}{*}{Vehicle}}    & PETR           & 0.497          & 0.704          & 0.156          & 0.126          & 0.584                  & 0.074          & 1.250          & 0.228          & 0.764          & 0.205                  \\
\multicolumn{1}{c|}{}                            & \textbf{+ours} & \textbf{0.508} & \textbf{0.684} & \textbf{0.153} & \textbf{0.114} & \textbf{0.596(+0.012)} & \textbf{0.435} & \textbf{0.769} & \textbf{0.162} & \textbf{0.205} & \textbf{0.528(+0.323)} \\ \midrule
\multicolumn{1}{c|}{\multirow{2}{*}{Pedestrian}} & PETR           & 0.457          & 0.776          & 0.266          & 0.859          & 0.411                  & 0.069          & 1.173          & 0.287          & 1.454          & 0.153                  \\
\multicolumn{1}{c|}{}                            & \textbf{+ours} & \textbf{0.492} & \textbf{0.715} & \textbf{0.262} & \textbf{0.606} & \textbf{0.482(+0.071)} & \textbf{0.401} & \textbf{0.812} & \textbf{0.270}  & \textbf{1.087} & \textbf{0.353(+0.200)} \\ \midrule
\multicolumn{1}{c|}{\multirow{2}{*}{Cyclist}}    & PETR           & 0.318          & 0.713          & 0.229          & 0.569          & 0.407                  & 0.000          & 1.288          & 0.542          & 1.401          & 0.077                  \\
\multicolumn{1}{c|}{}                            & \textbf{+ours} & \textbf{0.328} & \textbf{0.668} & \textbf{0.209} & \textbf{0.465} & \textbf{0.440(+0.033)} & \textbf{0.154} & \textbf{0.799} & \textbf{0.258} & \textbf{0.854} & \textbf{0.259(+0.182)} \\ \midrule
\multicolumn{1}{c|}{\multirow{2}{*}{all}}        & PETR           & 0.424          & 0.731          & 0.217          & 0.518          & 0.467                  & 0.048          & 1.237          & 0.352          & 1.206          & 0.132                  \\
\multicolumn{1}{c|}{}                            & \textbf{+ours} & \textbf{0.443} & \textbf{0.689} & \textbf{0.208} & \textbf{0.395} & \textbf{0.506(+0.039)} & \textbf{0.330} & \textbf{0.793} & \textbf{0.230}  & \textbf{0.715} & \textbf{0.375(+0.243)} \\ \bottomrule
\end{tabular}
\label{table_waymo_petr_full+sparse}
\end{table*}

\begin{table*}[]
\caption{Ablation study of object-scene-camera decomposition and recomposition operations.}
\centering
\begin{tabular}{@{}c|c|c|ccc|ccc@{}}
\toprule
\multirow{2}{*}{\textbf{Object-Scene Recomposition}} & \multirow{2}{*}{\textbf{Object-Scene Decomposition}} & \multirow{2}{*}{\textbf{Camera Pose Pertubation}} & \multicolumn{3}{c|}{$\ AP_{3D}$}        & \multicolumn{3}{c}{$\ AP_{BEV}$}        \\
                                        &                                         &                                                 & easy  & mod   & hard  & easy  & mod   & hard  \\ \midrule
-                                       & -                                       & -                                               & 15.89          & 12.76          & 10.57          & 23.14          & 18.17          & 15.31          \\
\checkmark               & -                                       & -                                               & 20.69          & 16.92          & 14.34          & 27.95          & 21.96          & 19.22          \\
\checkmark               & \checkmark               & -                                               & 23.30          & 19.02          & 16.35          & 32.00          & 25.12          & 21.99          \\
\textbf{\checkmark}               & \textbf{\checkmark}               & \textbf{\checkmark}                       & \textbf{26.57} & \textbf{20.42} & \textbf{17.34} & \textbf{36.37} & \textbf{26.74} & \textbf{23.07} \\ \bottomrule
\end{tabular}
\label{table_rde_ablation}
\end{table*}

\begin{table}[]
\centering
\caption{Ablation study of different decomposition ratios.}
\begin{tabular}{@{}c|ccc|ccc@{}}
\toprule
\multirow{2}{*}{\textbf{\begin{tabular}[c]{@{}c@{}}Decomposition \\ Ratio(\%)\end{tabular}}} & \multicolumn{3}{c|}{$\ AP_{3D}$}        & \multicolumn{3}{c}{$\ AP_{BEV}$}        \\
                                                                                             & easy  & mod   & hard  & easy  & mod   & hard  \\ \midrule
0                                                                                            & 15.89          & 12.76          & 10.57          & 23.14          & 18.17          & 15.31          \\
25                                                                                           & 18.24          & 14.96          & 12.64          & 24.31          & 19.88          & 17.35          \\
50                                                                                           & 19.92          & 15.71          & 14.01          & 27.31          & 21.80          & 18.94          \\
\textbf{100}                                                                                          & \textbf{22.44} & \textbf{17.97} & \textbf{15.28} & \textbf{31.94} & \textbf{24.59} & \textbf{21.70} \\ \bottomrule
\end{tabular}
\label{table_decomposition_ablation}
\end{table}

\subsubsection{Fully-Supervised Setting}
We evaluate on both the test set and the validation split on KITTI dataset. 
Table \ref{table_baseline_cmp_pro} shows that our method significantly improves the accuracy of five base models MonoDLE~\citep{ma2021delving}, GUPNet~\citep{lu2021geometry}, DID-M3D~\citep{peng2022did}, MonoDETR~\citep{zhang2023monodetr}, \IJCVRevised{and MonoDGP}~\citep{pu2025monodgp}. 
Specifically, the results of moderate-level of $AP_{3D}$ of the five base models are improved by 5.81, 4.33, 4.16, 4.39, and 2.38 
(48\%, 29\%, 26\%, 27\%, and 13\% relative improvement)
on the test set, and 4.62, 3.51, 4.35, 2.82, 1.56 on the validation split.
We attribute the significant improvements in these base models to the ability of our scheme to more efficiently leverage the knowledge in limited training data.

Table \ref{table_kitti_testset_pro} shows the comparison to SOTA models in recent years on the KITTI test set. Especially for NeurOCS ~\citep{min2023neurocs} and MonoLSS ~\citep{li2024monolss}, which lead easy-level and moderate/hard-level performance, our scheme improves all three levels of $AP_{3D}$ and $AP_{BEV}$ by 0.8, 1.95, 0.91 and 2.41, 0.87, 0.54, respectively. Our method thereby establishes a new SOTA for M3OD.
Notably, our scheme holds great potential to further boost performance with stronger base models upon codes released in the future.

\subsubsection{Sparsely-Supervised Setting}
We sparsely annotate all 96 clips in the KITTI train split and randomly select a few clips to fully annotate. We choose three different annotation ratios, i.e., 10\%, 20\% and 50\%, for experiment.
Figure \ref{fig_label_ratio} shows the results of these three annotation ratios on MonoDLE, GUPNet, DID-M3D, and MonoDETR. Our scheme with fewer sparse annotations, especially 10\% for MonoDLE, GUPNet and DID-M3D, achieves on-par performance with the fully-supervised setting of these base models. The performance increases with the annotation ratio. 

Notably, the performance for 100\% annotation in this setting is higher than in the fully-supervised setting because the model is pretrained first. 
Table \ref{table_10label_4bsl} further shows the results of our scheme on the three base models under the sparsely-supervised setting with a lower annotation ratio. Our scheme, with fewer annotations (10\% for the first three base models), can achieve on-par performance with the fully-supervised training of these three models in the KITTI train split. When both our scheme and the base models use the same sparse annotations, the performance of the base models drops significantly, while our scheme achieves significant gains compared to the base models.
This demonstrates the strong data exploitation ability of our method.

\subsection{Main Results on \IJCVRevised{Waymo-Mono}}
We further evaluate on the more complicated Waymo Dataset. We deploy our scheme on two representative types of base models, i.e., the convolution-based DID-M3D ~\citep{peng2022did} and the transformer-based MonoDETR ~\citep{zhang2023monodetr}. Table \ref{table_kitti_testset_pro} showed that our scheme implemented on these two models achieves SOTA on KITTI. Table \ref{table_waymo_did_detr_3cls_res} shows the experimental results. We can see that, although the dataset scale of Waymo is 10$\times$ larger than KITTI, we also achieve a significant improvement to the base models over three classes. This indicates that, not only the small-scale dataset but also the large-scale dataset suffers from the tight entanglement issue. Although a larger dataset can alleviate this issue, due to the efficient data mining capabilities of our scheme, we can more efficiently utilize the training data to improve the performance for all object classes. Further, as shown in Table \ref{table_waymo_10_label_did_detr_car}, with the sparsely-supervised setting, only 10\% annotation is needed to achieve comparable performance with the fully-supervised setting. It also demonstrates the strong data exploitation ability of our scheme.

\IJCVRevised{Moreover, Table \ref{table_waymo_did_distance} shows the performance of our scheme based on DID-M3D across different distances under both fully-supervised and sparsely-supervised settings. The experiment shows that our scheme effectively improves the detection performance within 0-30m and 30-50m ranges. For objects in the 50-$\infty$m range, considering the ill-posed nature of M3OD, objects at such a far distance are hard to detect, so our scheme shows limited improvement. }

\subsection{\IJCVRevised{Main Results on Waymo-Ring}}
\IJCVRevised{To evaluate the applicability of our scheme on the multi-camera Waymo-Ring dataset, we deploy it on the representative multi-camera 3D object detector PETR~\citep{liu2022petr}. Table \ref{table_waymo_petr_full+sparse} shows the experiment results. We can see that, although under the more challenge multi-camera setting, our scheme still effectively improves the detection performance across all categories of PETR, under both fully-supervised and sparsely-supervised settings. This indicates that the tight entanglement issue still occur in multi-camera setting and our scheme can alleviate it.}

\begin{table}[]
\caption{
Ablation study of different counts of recomposition per scene.
}
\centering
\begin{tabular}{@{}c|ccc|ccc@{}}
\toprule
\multirow{2}{*}{\textbf{\begin{tabular}[c]{@{}c@{}}Counts of\\ Recomposition\end{tabular}}} & \multicolumn{3}{c|}{\textbf{$\ AP_{3D}$}}        & \multicolumn{3}{c}{\textbf{$\ AP_{BEV}$}}        \\
                                                                                        & easy  & mod   & hard  & easy  & mod   & hard  \\ \midrule
0                                                                                       & 15.89          & 12.76          & 10.57          & 23.14          & 18.17          & 15.31          \\
2                                                                                       & 18.89          & 14.86          & 12.51          & 26.67          & 20.44          & 17.40          \\
50                                                                                      & 21.11          & 17.00          & 14.47          & 29.17          & 22.86          & 19.78          \\
\textbf{200}                                                                                     & \textbf{23.30} & \textbf{19.02} & \textbf{16.35} & \textbf{32.00} & \textbf{25.12} & \textbf{21.99} \\ \bottomrule
\end{tabular}
\label{table_recomposition_ablation}
\end{table}

\begin{table}[]
\caption{Ablation study of scene utilization before and after applying our scheme.}
\centering
\begin{tabular}{@{}c|c@{}}
\toprule
\textbf{Method}         & Object-scene composition per scene \\ \midrule
plain          & $\ \approx$4             \\
\textbf{+ours} & \textbf{$\ \approx$700}  \\ \bottomrule
\end{tabular}
\label{table_scene_utilization}
\end{table}

\subsection{Ablation Study}
We investigate the effect of each component in our data manipulation scheme on the KITTI validation split using MonoDLE.

\subsubsection{Effectiveness of Object-Scene-Camera Decomposition and Recomposition}
In our scheme, we propose three strategies to alleviate the tight entanglement of the three intenties, i.e., object, scene, and camera pose, in training data. 
More specifically, object-scene decomposition helps the network alleviate the overfitting to uniform training data, object-scene recomposition helps explore more object-scene and object-object relationships, and camera pose pertubation can extend the camera pose of a frame from fixed value to diverse variations.
Table \ref{table_rde_ablation} validates the effectiveness of the combination of the three proposed strategies. We can see that each of these three components contributes significantly to the performance. By combining all components, we achieve a significant gain in model performance.

\subsubsection{Effectiveness of Decomposition Operation}
We validate the effectiveness of our decomposition operation to alleviate the overfitting issue to the uniform training data. We randomly selected 25\% clips, 50\% clips and all data from train split to apply decomposition and obtain different scales of empty scenes and an object database. For the empty scenes, we randomly select objects from the database and insert them into the position of the original objects in the scene. Table \ref{table_decomposition_ablation} shows the results. We can see that, as the proportion of decomposed scenes increases, the performance of the model continues to improve. This indicates that the model tends to overfit on uniform data and our decomposition operation can alleviate it.

\begin{figure*}[]
\centering
\includegraphics[width=0.95\textwidth]{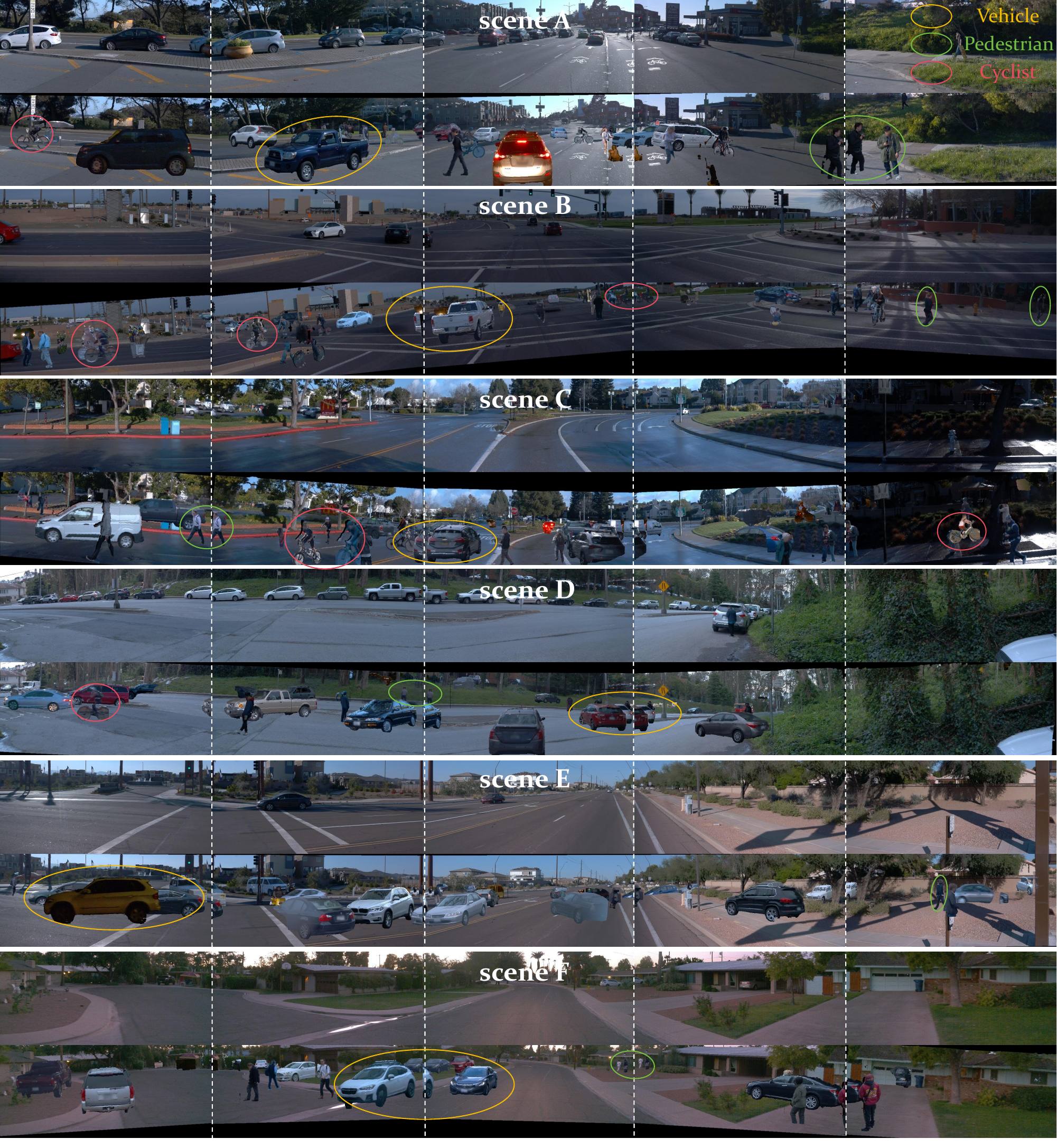} 
\caption{\IJCVRevised{Visualization of recomposed data based on decomposed data on Waymo-Ring. Our scheme can generate diverse data with randomly sampled object-scene-camera combinations online. For each scene, the top row shows the raw scene, and the bottom row shows the recomposed scene. We circle some of the recomposed samples of each category in the figure with different color.}}
\label{fig_recom_vis}
\end{figure*}

\begin{table}[]
\caption{Ablation study of different degrees of camera pose (pitch and roll) perturbation.}
\centering
\begin{tabular}{@{}c|ccc|ccc@{}}
\toprule
\multirow{2}{*}{\textbf{$\theta$ \& $\alpha$ (°)}} & \multicolumn{3}{c|}{\textbf{$AP_{3D}$}}        & \multicolumn{3}{c}{\textbf{$AP_{BEV}$}}        \\
                                                       & easy  & mod   & hard  & easy  & mod   & hard  \\ \midrule
0                                                      & 15.89          & 12.76          & 10.57          & 23.14          & 18.17          & 15.31          \\
1                                                      & 20.84          & \textbf{16.86} & 13.50          & 28.07          & \textbf{21.99} & 18.11          \\
\textbf{2}                                             & \textbf{21.76} & 16.13          & \textbf{13.60} & \textbf{28.71} & 21.30          & \textbf{18.41} \\
3                                                      & 19.98          & 15.48          & 12.90          & 27.98          & 21.03          & 17.79          \\ \bottomrule
\end{tabular}
\label{table_pitch_roll_ablation}
\end{table}
\begin{table}[]
\caption{Ablation study of different degrees of camera pose (translation) perturbation.}
\centering
\begin{tabular}{@{}c|ccc|ccc@{}}
\toprule
\multirow{2}{*}{\textbf{$\ \epsilon_{z}$(m)}} & \multicolumn{3}{c|}{\textbf{$\ AP_{3D}$}}        & \multicolumn{3}{c}{\textbf{$\ AP_{BEV}$}}        \\
                                              & easy  & mod   & hard  & easy  & mod   & hard  \\ \midrule
0                                             & 15.89          & 12.76          & 10.57          & 23.14          & 18.17          & 15.31          \\
1                                             & 19.90          & 15.40          & 12.93          & 27.98          & 21.13          & 18.09          \\
\textbf{2}                                    & \textbf{21.21} & \textbf{16.20} & \textbf{14.10} & \textbf{28.85} & \textbf{22.04} & \textbf{18.97} \\
3                                             & 19.22          & 15.16          & 12.76          & 27.38          & 21.62          & 18.75          \\ \bottomrule
\end{tabular}
\label{table_translation_ablation}
\end{table}

\subsubsection{Effectiveness of Recomposition Operation}
We validate the effectiveness of our recomposition operation to alleviate the underutilization issue of the object-scene relationship. We apply different numbers of recompositions for each scene while keeping the total training epochs the same. Table \ref{table_recomposition_ablation} shows the results. We can see that, as the number of recompositiosn increases, more variance of appearance and 2D size of objects and more diverse combinations of objects and scenes are introduced. This allows the model to exploit more information and achieve higher performance.

Table \ref{table_scene_utilization} also reflects how our method greatly utilizes the object-scene relationship. For the original data in the KITTI train split, each scene contains, on average, only 4 object-scene compositions. With our scheme, after 200 iterations, we generate up to approximately 700 object-scene compositions per scene.

\begin{figure*}[]
\centering
\includegraphics[width=0.95\textwidth]{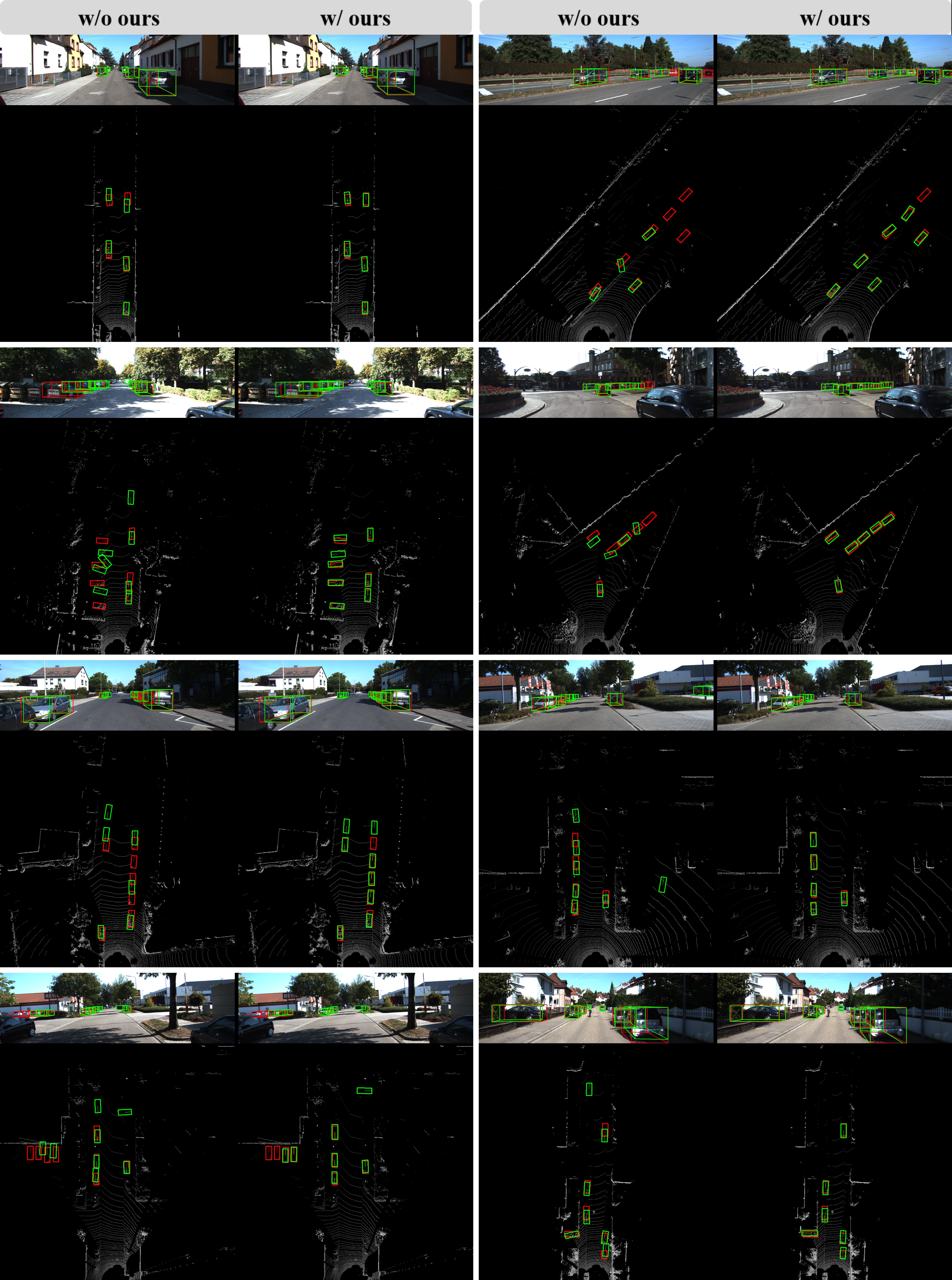} 
\caption{Qualitative comparison of detection results on the KITTI validation set for DID-M3D training with and without our scheme.
For each scene, the left and right images show the results without and with our scheme, respectively. Ground truth boxes are shown in red, and predicted boxes are shown in green.}
\label{fig_qualitative_results}
\end{figure*}

\subsubsection{Effectiveness of Camera Pose Perturbation}
We validate the effectiveness of our camera pose perturbation operation to alleviate the limited camera pose variation issue. Tables \ref{table_pitch_roll_ablation} and \ref{table_translation_ablation} show the effect of applying our camera pose perturbation operation on the performance of the base model. After applying pitch and roll perturbation, the network learns an extrinsic-invariant representation against pose variation. By applying translation perturbation, we can dynamically increase the temporal resolution, enabling  the network to learn from the data effectively.

\begin{table}[]
\caption{Ablation study of different ratios of empty scenes per batch.}
\centering
\begin{tabular}{@{}c|ccc|ccc@{}}
\toprule
\multirow{2}{*}{\textbf{$\ r_{empty}$}} & \multicolumn{3}{c|}{$\ AP_{3D}$}        & \multicolumn{3}{c}{$\ AP_{BEV}$}        \\
                                        & easy  & mod   & hard  & easy  & mod   & hard  \\ \midrule
1.0                                     & 19.83          & 13.65          & 10.14          & 27.55          & 19.13          & 14.76          \\
\textbf{0.5}                                     & \textbf{26.57} & \textbf{20.42} & \textbf{17.34} & \textbf{36.37} & \textbf{26.74} & \textbf{23.07} \\
0                                       & 23.07          & 17.92          & 15.33          & 31.58          & 23.68          & 20.78          \\ \bottomrule
\end{tabular}
\label{table_mix_sampling}
\end{table}

\begin{table}[]
\caption{Comparison with copy-paste strategies.}
\centering
\begin{tabular}{@{}c|ccc@{}}
\toprule
\multirow{2}{*}{\textbf{Model}}                                                                         & \multicolumn{3}{c}{\textbf{$\ AP_{3D} / AP_{BEV}$}}                      \\
                                                                                                        & easy                   & mod                    & hard                   \\ \midrule
\begin{tabular}[c]{@{}c@{}}MonoDLE\\ ~\citep{ma2021delving}\end{tabular}          & 17.45 / 24.97          & 13.66 / 19.33          & 11.68 / 17.01          \\ \midrule
\begin{tabular}[c]{@{}c@{}}+EGC\\ ~\citep{lian2022exploring}\end{tabular}         & 22.79 / 30.44          & 16.74 / 22.71          & 14.60 / 19.60          \\ \midrule
\begin{tabular}[c]{@{}c@{}}+MonoSample\\ ~\citep{qiao2024monosample}\end{tabular} & 19.71 / 26.66          & 15.82 / 21.14          & 13.58 / 18.36          \\ \midrule
\begin{tabular}[c]{@{}c@{}}\IJCVRevised{+MonoPlace3D}\\ ~\citep{parihar2025monoplace3d}\end{tabular} & 22.49 / -          & 15.44 / -          & 12.89 / -          \\ \midrule
\textbf{+ours}                                                                                          & \textbf{26.87 / 35.52} & \textbf{19.48 / 25.07} & \textbf{16.02 / 21.15} \\ \bottomrule
\end{tabular}
\label{table_copy_paste_comparision}
\end{table}

\subsubsection{Mix sampling of raw/empty scenes in training}
Table \ref{table_mix_sampling} shows a comparison of different mixing ratios between raw scenes and empty scenes. When only raw scenes are sampled, recomposition flexibility is poor due to fixed object semantics. As a result, it is difficult to fully utilize object-scene relationship, and leading to network overfitting. However, if only empty scenes are sampled, recomposition flexibility is maximized. Nevertheless, performance remains poor due to the domain gap between synthetic and real data. To make a trade-off, we mix raw scenes and empty scenes with the ratio 0.5. With this mixed sampling strategy, the network can train stably while benefiting from data recomposition.

\subsubsection{Comparison with Copy-Paste strategies}
We also compare our scheme with \IJCVRevised{three} representative copy-paste strategies in EGC ~\citep{lian2022exploring}, MonoSample ~\citep{qiao2024monosample}, \IJCVRevised{and MonoPlace3D} ~\citep{parihar2025monoplace3d}. Table  \ref{table_copy_paste_comparision} shows the results. We can see that our scheme holds huge performance gains compared to these three methods. The reason is that, although copy-paste methods partly alleviate the tight entanglement issue to some degree, e.g., the object-scene
relationship, but cannot alleviate the two issues of overfitting to uniform training data and limited camera pose variation.
For comparison, we delve into the tight entanglement issue and propose three strategies to fully mitigate it. Further, we can achieve 2D-3D geometric consistency after data manipulation.

\subsubsection{\IJCVRevised{Quantitative results of recomposed images}}
\IJCVRevised{We use FID and LPIPS to evaluate the quantitative quality of recomposed images in our scheme. The experiment is conducted on the KITTI train split. Table \ref{table_fid_lpips} shows the FID and LPIPS scores of our results. We achieve better results than the simple 2D copy-paste because we consider the geometric consistency.}  

\begin{table}[]
\caption{\IJCVRevised{Quantitative results of recomposed images.}}
\centering
\begin{tabular}{c|cc}
\hline
\textbf{Method}        & FID $\downarrow$            & LPIPS $\downarrow$          \\ \hline
2D copy paste & 9.977          & 0.098          \\
\textbf{ours} & \textbf{8.757} & \textbf{0.068} \\ \hline
\end{tabular}
\label{table_fid_lpips}
\end{table}

\begin{table}[]
\caption{\IJCVRevised{Computational costs of our scheme.}}
\centering
\begin{tabular}{@{}c|c|c@{}}
\toprule
\textbf{Type}                                       & \textbf{Action}                       & \textbf{Speed}   \\ \midrule
\multirow{2}{*}{Offline Data Manipulation} & Scene Database Construction  & 3.5h    \\
                                           & Object Database Construction & 1h      \\ \midrule
\multirow{2}{*}{Online Data Manipulation}  & Object-Scene Recomposition   & 5fps    \\
                                           & Camera Pose Perturbation     & 2500fps \\ \midrule
\multirow{4}{*}{Training}                  & DID-M3D                      & 24h     \\
                                           & DID-M3D+ours                 & 40h     \\
                                           & PETR                         & 12.5h   \\
                                           & PETR+ours                    & 12.5h   \\ \bottomrule
\end{tabular}
\label{table_ops_speed}
\end{table}

\subsubsection{\IJCVRevised{Computational and storage costs}}
\IJCVRevised{
To evaluate the efficiency of our scheme, we show the computational and storage costs. Table \ref{table_ops_speed} shows the computation cost. Specifically, we evaluate the offline data manipulation, including scene database construction and object database construction on Waymo-Mono. The results show that the database construction cost is low thanks to our design. 

For the online data manipulation, we implement object-scene recomposition within dataloader sub-process on CPU, and we implement camera pose perturbation in the main-process on GPU. The results show that the camera pose perturbation is very efficient because all operations are implemented via matrix calculations,  which are well-suited for GPU. The object-scene recomposition is relatively slow because the calculation need to processed by the CPU step by step, but 5 fps is still a fast speed compared to the high-cost GAN or Diffusion schemes. 

Moreover, we show the training time of DID-M3D on Waymo-Mono and PETR on Waymo-Ring. The results show that our scheme has a certain degree of influence on DID-M3D and no influence on PETR. This is because, although the main online computation of our scheme is object-scene recomposition, it is implemented in the dataloader sub-process, which can run in parallel with network training. The training time of DID-M3D slows down because it is a lightweight network and the training speed bottleneck lies in the data pre-process running on CPU.

Finally, regarding the storage costs, for the Waymo-Ring dataset, additional storage required for the object database and scene database is 1.5GB and 97GB, compared to the raw images, which occupy 31GB.
}

\subsection{Qualitative Results}

\subsubsection{Visualization of Recomposed Data}
Figure \ref{fig_recom_vis} shows some recomposed data based on the combination of different categories of objects, scenes, and camera poses. Our scheme can explore their combination to generate diverse data online for efficient network training. We did not introduce complicated shadow rendering, and it does not hinder the performance enhancement.

\subsubsection{Visualization of Detection Results}
Figure \ref{fig_qualitative_results} shows some qualitative comparison of detection results for DID-M3D training with and without our scheme. We can see that,  the model trained without our scheme results in bad detection results. The reason lies that it suffers from the tight entanglement issue on uniform training data. After applying our scheme, it can alleviate the tight entanglement issue and well boost the performance of the model. 

\section{Conclusion}
In this paper, we observe that the crucial tight entanglement problem of the three independent entities, i.e., object, scene and camera pose, induces the challenging issues of insufficient utilization and overfitting to uniform training data for deep learning based monocular 3D object detection. To mitigate, we propose an online object-scene-camera decomposition and recomposition data manipulation scheme to more efficiently exploit training data. 
This scheme serves as a plug-and-play component to significantly boost existing M3OD models, which can work flexibly with both fully and sparsely supervised settings. Notably, our scheme holds great potential to achieve higher accuracy with stronger base models upon their codes released.
In the future, we plan to extend our scheme to \IJCVRevised{unsupervised setting}.

\noindent \textbf{Data Availability} The KITTI ~\citep{geiger2012we} and Waymo ~\citep{sun2020scalability} datasets used in this manuscript are deposited in publicly available repositories respectively: \href{https://www.cvlibs.net/datasets/kitti/}{\textcolor{blue}{https://www.cvlibs.net/datasets/kitti/}} and \href{https://waymo.com/open/data/perception}{\textcolor{blue}{https://waymo.com/open/data/perception}}

% Non-BibTeX users please use
{\footnotesize
	\bibliographystyle{plainnat}
	\bibliography{egbib}
}
% \end{flushend}
\balance

\end{document}